\ifdefined\pdfminorversion\pdfminorversion=7\fi
\documentclass[]{jingdong}
\usepackage{tabularx}
\usepackage{mathtools}
\usepackage{amssymb}
\usepackage{needspace}

\crefname{figure}{Fig.}{Figs.}
\Crefname{figure}{Fig.}{Figs.}
\crefname{table}{Table}{Tables}
\Crefname{table}{Table}{Tables}

\AddToHook{cmd/section/before}{\Needspace*{5\baselineskip}}
\AddToHook{cmd/subsection/before}{\Needspace*{3\baselineskip}}
\AddToHook{cmd/subsubsection/before}{\Needspace*{3\baselineskip}}

\newcommand{\paperfigure}[1]{\includegraphics[width=\linewidth]{fig/#1.pdf}}
\usepackage{tikz}
\usetikzlibrary{arrows.meta,positioning,fit,calc}
\newcommand{\resourcecodeicon}{%
  \raisebox{-0.05em}{\includegraphics[height=0.84em]{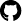}}}
\newcommand{\resourcedataicon}{%
  \raisebox{-0.18\height}{\includegraphics[height=1.02em]{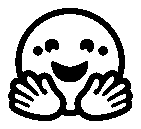}}}
\newcommand{\resourcewebicon}{%
  \raisebox{-0.07em}{\includegraphics[height=0.86em]{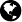}}}
\newcommand{\resourcelink}[3]{\href{#1}{\mbox{\smash{\textcolor{black}{#2}}\hspace{0.4em}#3}}}

\newcommand{\bench}{ROAM-Bench}

\newcommand{\ind}{\mathbb{1}}
\newcommand{\ans}{\mathrm{A}}
\newcommand{\nec}{\mathrm{N}}
\newcommand{\suf}{\mathrm{S}}
\newcommand{\str}{\mathrm{T}}
\newcolumntype{L}[1]{>{\raggedright\arraybackslash}p{#1}}
\newcolumntype{Y}{>{\raggedright\arraybackslash}X}
\tikzset{
  paperbox/.style={draw,rounded corners=2pt,align=center,font=\small,inner sep=5pt},
  paperarrow/.style={-{Latex[length=2mm]},semithick},
  auditbox/.style={paperbox,fill=blue!5},
  agentbox/.style={paperbox,fill=green!5},
  warningbox/.style={paperbox,fill=orange!8}
}

\usepackage{colortbl}
\definecolor{tabfirst}{HTML}{A12D3A}
\definecolor{tabsecond}{HTML}{246296}
\definecolor{tabthird}{HTML}{557563}
\newcommand{\ranktag}[1]{\makebox[0pt][l]{\textsuperscript{\tiny\normalfont\,#1}}}
\newcommand{\rankone}[1]{\textcolor{tabfirst}{\textbf{#1}\ranktag{1}}}
\newcommand{\ranktwo}[1]{\textcolor{tabsecond}{#1\ranktag{2}}}
\newcommand{\rankthree}[1]{\textcolor{tabthird}{#1\ranktag{3}}}

\newcommand{\rankup}{{\normalsize$\uparrow$}}
\newcommand{\rankdown}{{\normalsize$\downarrow$}}

\ifdefined\pdfgentounicode
  \input glyphtounicode.tex
\fi
\makeatletter
\newcommand{\roamutftokens}[1]{%
  \begingroup
  \let\do\@makeother\dospecials
  \catcode32=10
  \endlinechar=-1
  \expandafter\scantokens\expandafter{\detokenize{#1}}%
  \endgroup}
\makeatother

\title{ROAM-Bench: Evaluating Embodied Active Spatial Intelligence through Evidence Verification}
\hypersetup{
  pdftitle={ROAM-Bench: Evaluating Embodied Active Spatial Intelligence through Evidence Verification},
  pdfauthor={Xingjian Ran, Xiaoye Mo, Sihao Liu, Li Luo, Taiqiang Wu, Ji Zhao, Wenqi Shao, Hongyang Li, Jiaqi Wang, Bo Dai}
}

\author[1,4]{Xingjian Ran}
\author[2]{Xiaoye Mo}
\author[1]{Sihao Liu}
\author[1]{Li Luo}
\author[1,4]{Taiqiang Wu}
\author[3,4]{Ji Zhao}
\author[4]{Wenqi Shao}
\author[4,\ddagger]{Hongyang Li}
\author[4,\dagger]{Jiaqi Wang}
\author[1]{Bo Dai}

\affiliation[1]{The University of Hong Kong}
\affiliation[2]{University of Electronic Science and Technology of China}
\affiliation[3]{Shanghai Innovation Institute}
\affiliation[4]{JD.com}

\contribution[\ddagger]{Project leader}
\contribution[\dagger]{Corresponding author}

\resources{%
  {\fontsize{12}{14}\selectfont\sffamily\bfseries
   \resourcelink{https://github.com/rxjfighting/ROAM-Bench}{\resourcecodeicon}{Code}%
   \hspace{2em}%
   \resourcelink{https://huggingface.co/datasets/xjran/ROAM-Bench-scenes}{\resourcedataicon}{Scene Assets}%
   \hspace{2em}%
   \resourcelink{https://rxjfighting.github.io/ROAM-Bench/}{\resourcewebicon}{Project Page}}%
}

\abstract{
Embodied spatial intelligence requires agents to gather information through exploration and interaction, then use it to reason and act.
Existing benchmarks insufficiently distinguish correct outcomes from completion of these task requirements.
We introduce \bench{} to evaluate these abilities under robot motion and collision constraints.
Independent checks use simulation state and episode records to assess outcomes and task requirements, with stronger checks of geometry, coverage, or placement.
It comprises 250 manually validated tasks across 11 types in five groups and 27 indoor environments, covering object and room search, counting, geometric judgments, direction reasoning, and relocation.
Human evaluators achieve 84\% strict task success on 100 sampled tasks, with every correct outcome satisfying the required checks.
Across seven frontier vision-language models, we identify two complementary failure modes: correct outcomes with unmet task requirements, and incorrect outcomes despite passing stronger checks.
Oracle-input diagnostics show that more complete target observations and calibrated feedback improve final spatial judgments.
}

\begin{document}
\raggedbottom
\maketitle
\begin{figure}[t]
  \centering
  \includegraphics[width=\linewidth]{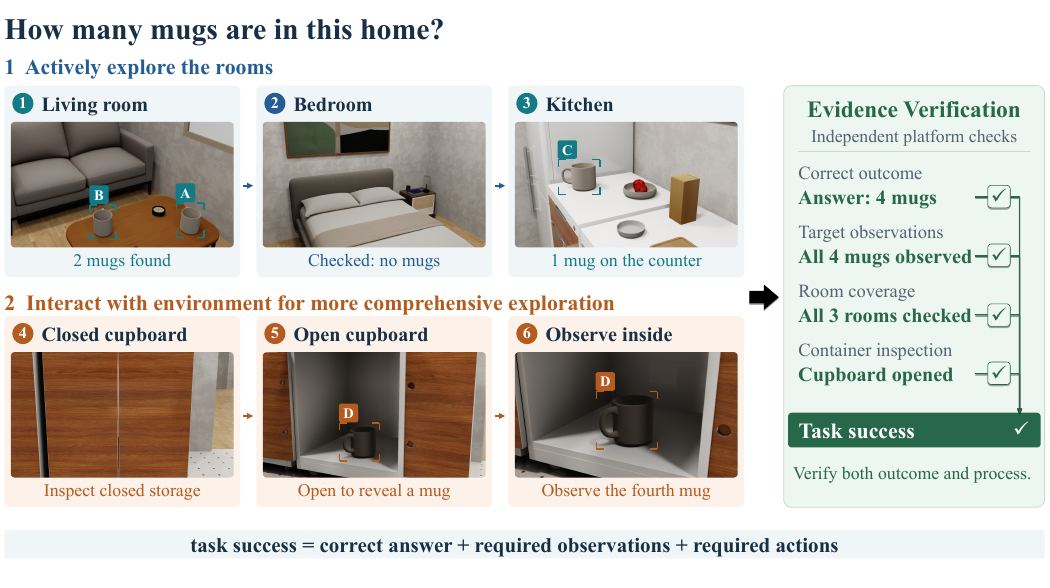}
  \caption{ROAM-Bench checks task outcomes and required observations and interactions. Counting mugs across a home requires searching multiple rooms, checking rooms without mugs, and opening a cupboard to find a hidden mug. Independent checks verify the final count, target observations, room coverage, and container inspection.}
  \label{fig:overview}
\end{figure}

\section{Introduction}
\label{sec:intro}

Embodied spatial intelligence requires agents to gather information through exploration and interaction. Counting mugs across a home, for example, requires searching multiple rooms, opening closed storage, and tracking objects across views to avoid double-counting (Fig.~\ref{fig:overview}). The agent must also choose views that reveal occluded targets and decide when the search is complete. Yet an agent may report the correct count without observing every mug or checking every room. It may also make incorrect spatial judgments after collecting the required observations. Outcome correctness must therefore be evaluated alongside the task's required observations and interactions.

Existing benchmarks address different parts of this process. VSI-Bench~\cite{yang2024thinking} and MMSI-Bench~\cite{yang2025mmsi} evaluate spatial question answering from preselected videos or images. EmbodiedQA~\cite{das2018embodied} and OpenEQA~\cite{majumdar2024openeqa} shift information gathering to the agent through question-driven navigation. EXPRESS-Bench~\cite{jiang2025express} assesses answer grounding in the final observed image, while ESI-Bench~\cite{hong2026esibench} evaluates perception, locomotion, and manipulation for spatial QA. EmbodiedBench~\cite{yang2025embodiedbench} and BEHAVIOR-Spatial~\cite{behavior2026spatial} evaluate embodied tasks and spatial skills. Across these settings, final scores do not consistently distinguish correct outcomes from completed observation and interaction requirements.

To address this gap, we introduce \bench{}, which evaluates task outcomes together with the observations and interactions required to reach them. The benchmark comprises 250 manually validated tasks across 11 types in five groups: grounding, aggregation, geometry, reference-frame integration, and goal-directed spatial action. Independent checks use simulation state and episode records to assess outcomes and separately verify required observations, search coverage, ordered interactions, geometry, and object placement. This verification helps distinguish missing observations from mistakes made after collecting them for deeper failure analysis. Task success requires these conditions together with a correct outcome, measuring how exploration, spatial judgment, and action work together.

Our platform, built on Isaac Sim~\cite{nvidiaIsaacSim51}, provides 27 furnished indoor environments generated with SceneSmith~\cite{pfaff2026scenesmith} and manually corrected for benchmark use. A vision-language model (VLM) chooses predefined navigation and interaction skills, including moving, observing, opening, picking, and placing. The simulator executes these atomic skills under declared motion and interaction constraints. After each skill, the agent receives a new view and action feedback before choosing its next step. The feedback reports collisions and object-handling outcomes, which the agent can use to adjust its plan.

Our contributions are \textbf{(1) A comprehensive benchmark and evaluation protocol for embodied spatial intelligence}, connecting exploration, spatial judgment, and goal-directed action. \textbf{(2) An executable exploration platform} with atomic actions, leveraging simulation oracles to verify task outcomes and task-required evidence. \textbf{(3) Human and frontier VLM evaluations with supporting diagnostics} covering information acquisition, precision, relocation failures, input sensitivity, and costs that include failures.

\section{Related Work}
\label{sec:related}

\begin{table}[t]
  \centering
  \small
  \DeclareRobustCommand{\settingyes}{\tikz[baseline=-1pt,x=1pt,y=1pt]{\draw[line width=0.8pt,line cap=round,line join=round] (0,2.5)--(2.3,0)--(7,6);}}
  \DeclareRobustCommand{\settingno}{\tikz[baseline=-0.5pt,x=1pt,y=1pt]{\draw[line width=0.7pt,line cap=round] (0,0)--(5,5) (0,5)--(5,0);}}
  \setlength{\tabcolsep}{4pt}
  \renewcommand{\arraystretch}{1.06}
  \caption{\settingyes / \settingno: included / excluded. \textsuperscript{*}: setting-dependent (OpenEQA: active navigation). Object actions: manipulation skills of scene objects, such as opening, picking, and placing. Body: robot motion/collision constraints. Evidence tiers: task-required evidence and stronger corroboration scored separately from outcome. Columns cover benchmark tasks.}
  \label{tab:benchmark_settings}
  \begin{tabularx}{\textwidth}{@{}L{0.235\textwidth}*{3}{>{\centering\arraybackslash}p{0.068\textwidth}}>{\centering\arraybackslash}p{0.095\textwidth}>{\centering\arraybackslash}p{0.080\textwidth}Y@{}}
    \toprule
    Benchmark & \shortstack{Spatial\\QA} & \shortstack{Active\\views} & \shortstack{Object\\actions} & \shortstack{Body\\constraints} & \shortstack{Evidence\\tiers} & Checks \\
    \midrule
    VSI-Bench~\cite{yang2024thinking} & \settingyes & \settingno & \settingno & \settingno & \settingno & Answer \\
    MMSI-Bench~\cite{yang2025mmsi} & \settingyes & \settingno & \settingno & \settingno & \settingno & Answer / reasoning \\
    OpenEQA~\cite{majumdar2024openeqa} & \settingyes & \settingyes & \settingno & \settingyes\textsuperscript{*} & \settingno & Answer / efficiency \\
    E3VS-Bench~\cite{sakamoto2026e3vs} & \settingyes & \settingyes & \settingno & \settingno & \settingno & Answer + views \\
    ESI-Bench~\cite{hong2026esibench} & \settingyes & \settingyes & \settingyes & \settingno & \settingno & Answer / views \\
    EmbodiedBench~\cite{yang2025embodiedbench} & \settingno & \settingyes & \settingyes & \settingyes\textsuperscript{*} & \settingno & Goal \\
    BEHAVIOR-Spatial~\cite{behavior2026spatial} & \settingyes & \settingyes & \settingyes & \settingyes\textsuperscript{*} & \settingno & QA / grounding / goals \\
    Embodied Agent Arena~\cite{huang2026arena} & \settingyes & \settingyes & \settingyes & \settingyes\textsuperscript{*} & \settingno & Outcome / conditions \\
    \textbf{\bench{}} & \settingyes & \settingyes & \settingyes & \settingyes & \settingyes & Outcome / evidence \\
    \bottomrule
  \end{tabularx}
\end{table}

\paragraph{Spatial intelligence and embodied task evaluation.}
Work in this area studies spatial understanding from images and video. VSI-Bench~\cite{yang2024thinking} evaluates video-based inference, and MindCube~\cite{wang2025mindcube} tests limited-view spatial modeling. MMSI-Bench~\cite{yang2025mmsi} evaluates multi-image reasoning with reference-assisted diagnosis. SpatialBench~\cite{xu2025spatialbench} and SITE~\cite{wang2025site} examine capability and modality coverage. BLINK~\cite{fu2024blink} tests depth and correspondence. Beyond interpreting visual observations, embodied evaluations test goal completion through navigation and interaction. EmbodiedBench~\cite{yang2025embodiedbench} evaluates vision-driven tasks, while Embodied Agent Arena~\cite{huang2026arena} checks outcomes and execution conditions. ALFRED~\cite{shridhar2019alfred} and BEHAVIOR-1K~\cite{li2024behavior} evaluate household activities. Room-to-Room~\cite{anderson2017vln} tests instruction-following navigation. Beyond Spatial Benchmarks~\cite{huang2026spatialnav} examines spatial-reasoning transfer. Habitat~\cite{savva2019habitat} and AI2-THOR~\cite{kolve2017ai2thor} provide simulation. EgoPlan-Bench2~\cite{qiu2024egoplan2} evaluates planning. BEHAVIOR-Spatial~\cite{behavior2026spatial} connects QA, grounding, and robot action.

\paragraph{Robot policies and high-level decisions.}
This line of work links visual and language understanding to robot behavior. SayCan~\cite{ahn2022saycan} grounds plans in skill affordances. PaLM-E~\cite{driess2023palme} integrates observations and language. RT-2~\cite{brohan2023rt2} transfers vision-language knowledge to control. OpenVLA~\cite{kim2024openvla} and Octo~\cite{octo2024policy} develop generalist policies. RoboVQA~\cite{sermanet2023robovqa} evaluates long-horizon reasoning.

\paragraph{Active embodied question answering.}
These tasks require agents to seek observations before answering. EmbodiedQA~\cite{das2018embodied} and Multi-Target EQA~\cite{yu2019multitarget} evaluate question-conditioned navigation. IQA~\cite{gordon2018iqa} tests information-revealing interactions. OpenEQA~\cite{majumdar2024openeqa} evaluates memory and exploration. Explore until Confident~\cite{ren2024explore} studies stopping, while EfficientEQA~\cite{cheng2024efficienteqa} combines exploration with retrieval. E3VS-Bench~\cite{sakamoto2026e3vs} evaluates viewpoint-dependent search. EXPRESS-Bench~\cite{jiang2025express} uses a VLM to score answer correctness and grounding in the final first-person image. ESI-Bench~\cite{hong2026esibench} tests perception, locomotion, and manipulation across 10 categories and compares active, passive, and oracle inputs.

\paragraph{From outcome scores to task completion assessment.}
These directions address complementary parts of embodied tasks: understanding space, choosing actions, and gathering observations. To evaluate how these capabilities work together under robot motion and collision constraints, \bench{} independently checks outcomes and task-required observations and interactions across complete episodes. Additional checks examine geometry and post-placement support. This connects final performance to the exploration and interaction process and supports detailed analysis of failures in observation acquisition, spatial judgment, and action (Table~\ref{tab:benchmark_settings}).
\section{Benchmark Design}
\label{sec:benchmark}

\subsection{Task Taxonomy and Interaction}
To evaluate embodied active spatial intelligence, we design tasks that combine exploration, spatial reasoning, and interaction in different ways. For example, counting may require searching several rooms, gap measurement requires observing both edges, and relocation requires finding and moving the correct object. On our platform based on Isaac Sim~\cite{nvidiaIsaacSim51}, the agent gathers observations and interacts with objects until it completes the task. We then check the final result and whether the task is carried out as required.

\begin{figure}[t]
  \centering
  \paperfigure{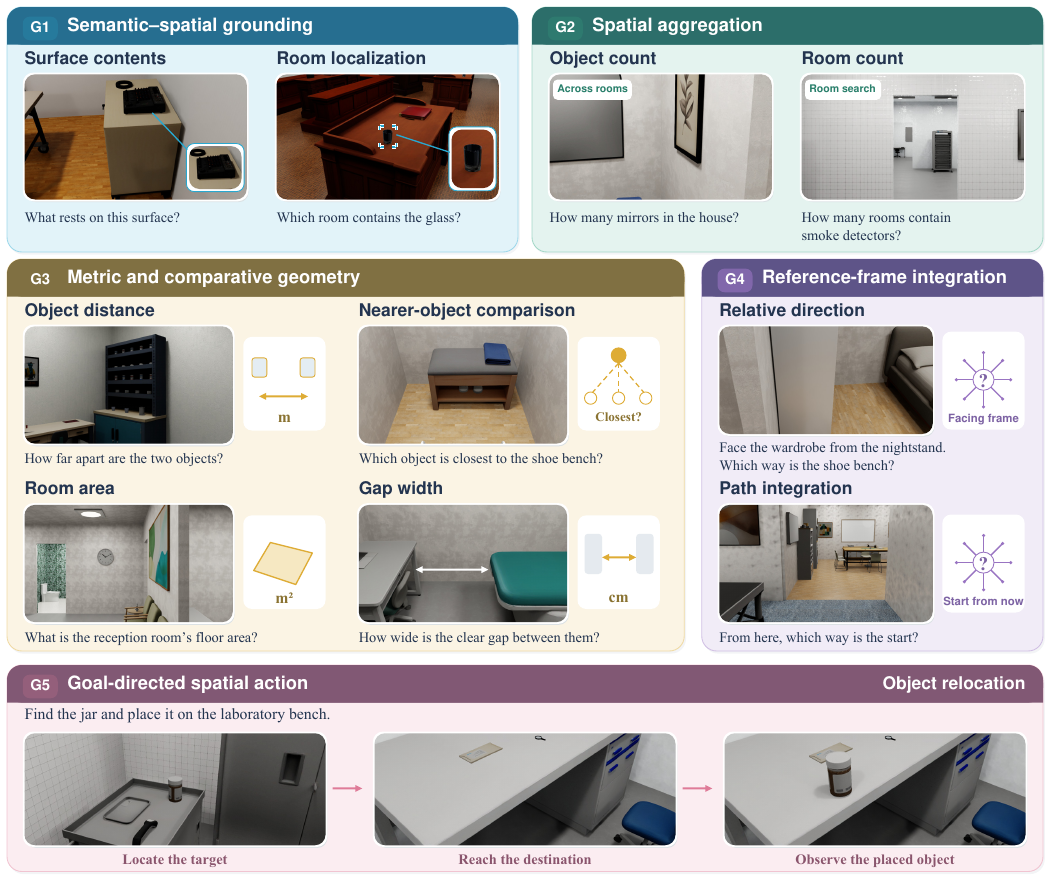}
  \caption{The 11 task types in five groups, illustrated with recorded camera observations, detail insets, and schematic annotations. Instructions are abbreviated. Relocation shows the target, destination, and placed-object stages of one episode.}
  \label{fig:task_atlas}
\end{figure}

We group the 11 task types by their main answer or action (Fig.~\ref{fig:task_atlas}). \textbf{G1, semantic-spatial grounding}, identifies objects and rooms. \textbf{G2, spatial aggregation}, counts objects or rooms across views. \textbf{G3, metric and comparative geometry}, estimates distances, areas, and gaps or compares proximity. \textbf{G4, reference-frame integration}, judges directions from reference objects or the agent's final position and heading. \textbf{G5, goal-directed spatial action}, finds and transfers the specified object to its destination.

Tasks describe targets through reference images of objects or surfaces, category names, and room names. To find a named room, the agent must infer its function from the layout and furnishings. The instructions also guide exploration and spatial judgment: counting questions specify where to search, while direction questions specify the position and orientation from which directions are judged. The agent therefore needs to interpret the task before deciding where to look and what to observe.

\paragraph{Atomic actions.}
All groups allow navigation---\emph{move, strafe, turn, look}---and object interaction---\emph{open, close, pick, place}. G1, G2, and G5 also permit navigation to an object or ground point selected in the latest image, emphasizing search and object handling. G3 and G4 require movement distances and directions to be specified, testing metric and reference-frame reasoning. To specify target objects and navigation locations without ambiguity, we use pixel selection in the latest camera image for object interaction and image-point navigation.

\subsection{Task Setup and Environments}
Each task specifies where the agent starts, what it must do, its action budget, and how its result and task execution are checked:
\begin{equation}
  \mathcal{T}_i = (s_i, x_i^0, p_i^0, q_i, g_i, B_i, b_i, v_i),
  \label{eq:task}
\end{equation}
The scene $s_i$ and initial state $x_i^0$ specify the objects, their geometry and poses, and interaction states such as whether containers are open. The starting pose $p_i^0$ gives the agent's position and heading. The remaining fields specify the instruction $q_i$, answer or goal rule $g_i$, evidence requirements $B_i$, action budget $b_i$, and environment and evaluation settings $v_i$. The required observation and interaction checks are derived from the instruction, including where the agent must search. They define N in Eq.~\eqref{eq:required_evidence} and are fixed together with the reference answers before evaluation.

\paragraph{Simulation environments.}
We use SceneSmith~\cite{pfaff2026scenesmith} to generate 27 furnished environments in five setting categories and manually correct them. Environments contain 1--13 rooms and 45--604 objects. Action budgets scale with room count to accommodate multi-room search and object transport. Category details and median scene sizes appear in Appendix~\ref{sec:appendix_tasks}.

\subsection{Task Generation and Validation}
We generate questions, reference answers, and goal conditions from the objects, room assignments, and geometry recorded in each scene. We check task feasibility through an oracle trajectory generated using simulator state and the available skills. We then manually inspect the scene to verify the instructions, answers or goals, and observation and interaction checks described below. This includes checking which objects and rooms must be counted, how directions are defined, and which object must be picked and subsequently placed at which destination.

\subsection{Outcome and Evidence Verification}
\label{sec:evidence}

\subsubsection{Agent Interaction}
\begin{figure}[t]
  \centering
  \includegraphics[width=\textwidth]{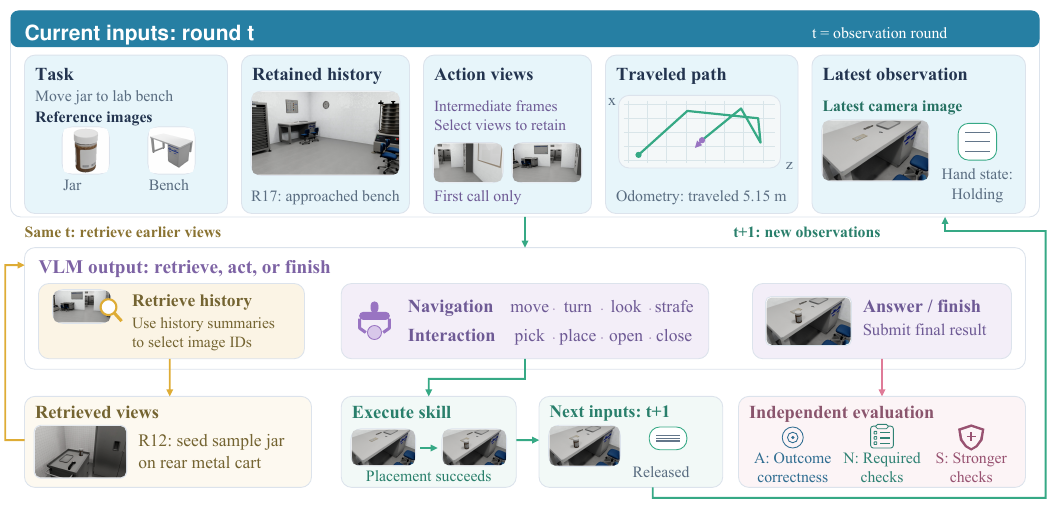}
  \caption{Agent interaction and independent evaluation. Retrieval supplements round $t$, a skill produces observations and feedback at $t+1$, and finishing triggers outcome and evidence checks. Saved episode images, a rendered reference object, and a path visualization illustrate the inputs and outputs.}
  \label{fig:protocol}
\end{figure}

At each round, the agent receives the task, reference images, latest camera image, available sensor readings, and feedback from its actions. Earlier observations and a visualization of the traveled path help it relate the current view to previous locations. Before executing a skill or submitting an answer or completion report, it can retrieve relevant earlier images using their text summaries. Retrieved images are added to the current round. Executing a skill produces the next observation and action feedback. Actions that select image points refer to the latest camera image. Once the agent finishes, the evaluator checks the final result and whether the task is carried out as required (Fig.~\ref{fig:protocol}).

\subsubsection{Outcome and Evidence Criteria}
\label{sec:evidence_criteria}
A correct answer alone does not show that the agent completes the task's required exploration and interactions. Conversely, an agent can complete the required observations yet answer incorrectly. We therefore check the result and task execution separately.

For task $i$, $\tau_i$ records its observations and actions, $y_i$ is the final response, and $x_i^{\mathrm{end}}$ is the final scene state. \emph{Outcome correctness} ($\ans$) checks whether the answer or final scene state meets the task's answer or goal rule. Numeric answers are converted to common units before scoring. Tolerances for distance $d$ and floor area $a$ are $\pm\max(2\,\mathrm{m},0.20d)$ and $\pm\max(5\,\mathrm{m}^2,0.25a)$. Gap-width tolerance is $\pm10\,\mathrm{cm}$. Human evaluators' RGB-only estimates pass the default tolerances in 110/120 judgments (91.7\%).

\emph{Task-required evidence} ($\nec$) checks whether the agent observes the required targets, searches the specified scope, and performs the required interactions. \emph{Stronger corroboration} ($\suf$) is designed to approximate sufficient evidence for answering the question or verifying that the goal is met. It requires all $\nec$ checks to pass, together with additional checks of geometry, viewing directions, or the released object's support at its destination. Thus, $\suf=1$ implies $\nec=1$. Both $\nec$ and $\suf$ are evaluated separately from $\ans$. We define
\begin{align}
  \ans_i &= \ind\{g_i(y_i,x_i^{\mathrm{end}})\text{ passes}\}, \\
  \nec_i &= \ind\{e_{\min}(\mathcal{T}_i,\tau_i)\text{ passes}\}, \label{eq:required_evidence} \\
  \suf_i &= \ind\{e_{\mathrm{strong}}(\mathcal{T}_i,\tau_i)\text{ passes}\}.
  \label{eq:receipts}
\end{align}
Here, $\mathcal{T}_i$ is the task specification, $g_i$ checks the answer or goal, and $e_{\min}$ and $e_{\mathrm{strong}}$ apply the $\nec$ and $\suf$ criteria, respectively. The indicator $\ind\{\cdot\}$ returns 1 when the check passes and 0 otherwise.

The primary metric is \emph{strict task success}:
\begin{equation}
  \str_i = \ans_i\nec_i.
  \label{eq:strict}
\end{equation}
Strict task success ($\str$) requires a correct result and completion of the required observations and interactions. Since $\ans_i$ and $\nec_i$ are binary, their product equals 1 only when both pass: $\str=\ans\text{ AND }\nec$. $\suf$ provides an additional diagnostic of the recorded episode.

\subsubsection{Task-Specific Evidence Requirements}
The checks follow each task's requirements. For example, counting checks which objects and rooms the agent observes, while relocation checks which object it picks and where it places it.

\paragraph{Observation and search.}
N checks whether the required targets appear in the $768\times432$ camera images. Counting also checks the specified object instances and search scope. A room counts as observed when the agent enters it or obtains an object view that passes the visibility check in that room. S can additionally require room visits and clear target views. Within a room, views from similar camera positions and orientations count once toward viewpoint coverage.

\paragraph{Geometry.}
For nearer-object questions, A checks the selected answer and N checks whether all referenced objects are observed. S additionally checks that depth-based distances identify the correct nearest candidate and distinguish it from the next closest after accounting for measurement uncertainty. Distance, direction, and gap-width tasks similarly compare geometry reconstructed from the recorded views with the task reference. For room area, N requires entering the target room, while S requires valid depth views from several directions.

\paragraph{Interaction.}
For relocation, N checks that the agent picks the specified object before placing that same object at the destination. For hidden targets, the checks also require opening the container before revealing the object. S additionally checks that the released object rests on the specified destination.

\paragraph{Landmarks and the answer frame.}
In path integration, the agent searches for two landmarks and reports where its starting position lies relative to its final position and heading. The provided start and final poses supply the information needed to judge this direction. Accordingly, $\ans$ checks the direction answer, while $\nec$ checks whether the agent observes both landmarks, regardless of order. $\suf$ additionally checks room and viewpoint coverage. A correct answer with $\nec=0$ therefore means that the required landmark exploration is unfinished.

\section{Experimental Evaluation}
\label{sec:protocol}
\label{sec:experiments}

\subsection{Evaluation Setup}
\label{sec:evaluation_setup}

\subsubsection{Models and Interaction}
We evaluate the seven VLMs in Table~\ref{tab:completed_models} on the same 250 tasks across 27 environments. Each episode starts from its defined scene state and camera pose. Agents use the public observation and skill interfaces until answering, signaling completion, or exhausting their room-scaled action budgets. High-level action decisions are charged.

Human evaluators complete a stratified sample of 100 tasks, one person per task, through the public observation and action interface. In a separate audit, human evaluators check N/S conditions for 200 episodes. Agreement with automatic checks is 97.4\% for N (196 ratable cases) and 93.3\% for S (194 ratable cases).

\subsubsection{Evaluation Metrics}
All models complete the same task set $\mathcal{C}$. We measure answer or goal accuracy and strict task success as
\begin{equation}
  \mathrm{Acc}_{\ans}=\frac{1}{|\mathcal{C}|}\sum_{i\in\mathcal{C}}\ans_i,
  \qquad
  \mathrm{Acc}_{\str}=\frac{1}{|\mathcal{C}|}\sum_{i\in\mathcal{C}}\str_i.
  \label{eq:accuracy}
\end{equation}
N and S are averaged over the same tasks. Human scores use the same rules on the 100-task sample. Error given S, $P(\ans=0\mid\suf=1)$, is the fraction of S-passing episodes with incorrect outcomes, using each model's S-passing episodes as the denominator.

The difference between correctness and strict task success is
\begin{equation}
  G=\frac{\sum_{i\in\mathcal{C}}\ans_i(1-\nec_i)}{|\mathcal{C}|}
   =\mathrm{Acc}_{\ans}-\mathrm{Acc}_{\str}.
  \label{eq:gap}
\end{equation}
$G$ is the \emph{correctness--task-completion gap}: the fraction of all episodes with $\ans=1$ and $\nec=0$. Table~\ref{tab:completed_models} reports the corresponding number of gap cases, $|\mathcal{C}|G$.

\subsection{Main Results}
\label{sec:main_results}

\begin{table}[t]
\centering\small
\setlength{\tabcolsep}{3pt}
\caption{Results on 250 tasks per VLM and a stratified 100-task human sample (one person per task). Bold T requires a correct outcome with task-required evidence (A AND N). ``Gap cases'' counts episodes with $\ans=1$ and $\nec=0$ and equals $|\mathcal{C}|G$. Error given S reports errors/S-passing episodes. Average actions includes failures. Superscripts \textcolor{tabfirst}{\textbf{1}}, \textcolor{tabsecond}{2}, and \textcolor{tabthird}{3} mark the top three VLM values per column; ties share a rank. Arrows indicate the ranking direction.}
\label{tab:completed_models}
\begin{tabular*}{\textwidth}{@{\extracolsep{\fill}}lc>{\bfseries}cccccc@{}}
\toprule
& \multicolumn{4}{c}{Outcome and evidence} & \multicolumn{2}{c}{Diagnostic gaps} & Cost \\
\cmidrule(lr){2-5}\cmidrule(lr){6-7}\cmidrule(l){8-8}
Model & A (\%) \rankup & T (\%) \rankup & N (\%) \rankup & S (\%) \rankup & Gap cases (n) \rankdown & Error given S (\%) \rankdown & Avg. actions \rankdown \\
\midrule
GPT-5.6-sol & 34.4 & 19.2 & 38.0 & 31.2 & 38 & 47.4 (37/78) & 74.11 \\
GPT-6.1-sol & \ranktwo{58.8} & \ranktwo{46.8} & \ranktwo{67.6} & \ranktwo{56.8} & \rankthree{30} & \ranktwo{31.0 (44/142)} & \ranktwo{50.82} \\
DeepSeek-V4.1-Flash & 19.2 & 10.4 & 32.0 & 23.6 & \ranktwo{22} & 72.9 (43/59) & 66.27 \\
Claude Sonnet 5.5 & 44.0 & 31.6 & 53.2 & 41.2 & 31 & 39.8 (41/103) & 60.30 \\
Kimi K3.1 Pro & 44.0 & 28.4 & 55.2 & 46.4 & 39 & 45.7 (53/116) & 79.10 \\
Claude Opus 5.5 & \rankthree{54.4} & \rankthree{40.0} & \rankthree{63.2} & \rankthree{52.8} & 36 & \rankthree{34.1 (45/132)} & \rankthree{53.18} \\
GPT-6-astra & \rankone{73.2} & \rankone{69.2} & \rankone{83.6} & \rankone{68.0} & \rankone{10} & \rankone{15.9 (27/170)} & \rankone{43.62} \\
\midrule
Human & 84.0 & 84.0 & 92.0 & 73.0 & 0 & 4.1 (3/73) & 53.55 \\
\bottomrule
\end{tabular*}
\end{table}

\begin{figure}[t]
  \centering
  \includegraphics[width=\linewidth]{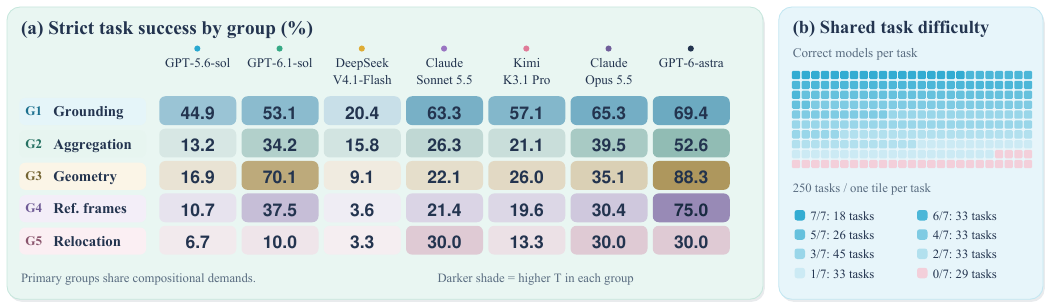}
  \caption{Task-group performance and shared task difficulty across seven models. Left: strict task success within each primary task group. Right: distribution of the 250 tasks by the number of models answering correctly. DeepSeek denotes DeepSeek-V4.1-Flash.}
  \label{fig:profiles_difficulty}
\end{figure}

Table~\ref{tab:completed_models} reports overall performance, and Figure~\ref{fig:profiles_difficulty} shows task-group scores and shared task difficulty. Human evaluators attain 84.0\% strict task success on their 100-task sample, demonstrating task feasibility through the public interface. Three findings characterize the results.

\paragraph{Strict task success remains below 70\%.}
GPT-6-astra achieves the highest strict task success, but still fails on nearly one-third of the tasks. The remaining models achieve strict task success on fewer than half of the tasks. Moreover, 29 tasks receive no correct answer from any model. Together, these results indicate persistent difficulty in connecting exploration, spatial judgment, and action.

\paragraph{VLMs struggle with both task coverage and spatial judgment.}
Correct outcomes do not always mean that required observations and interactions are complete. The 106 path-integration gaps concern unmet landmark exploration. Across the other task types, 100/1,540 episodes have correct outcomes but unmet task requirements. All seven models exhibit this gap (6--18 cases each).

Conversely, incorrect outcomes also occur after stronger corroboration checks (S) pass. Every VLM has a higher error rate in these episodes than human evaluators. In the 100-task human sample, all correct outcomes meet required coverage, and errors after S passes are rare. Together, these findings motivate linking target search and coverage tracking with reliable instance-level spatial judgment.

\paragraph{Task-group performance reveals uneven strengths.}
Relocation is the weakest group for six of the seven models, while locating objects and identifying their rooms is generally easier. GPT-6.1-sol performs better on geometry than on grounding, whereas both Claude models show the reverse pattern. These profiles distinguish scale estimation, cross-view instance tracking, and intended-object transfer.

\section{Analysis and Discussion}
\label{sec:analysis}

Errors after observation checks pass reveal spatial-judgment difficulties, while human evaluators identify incomplete target observation. Recorded answers, actions, and input comparisons locate frontier VLM bottlenecks.

\subsection{Spatial Judgment across Task Groups}
\label{sec:spatial_reasoning}
We examine G1--G4 using errors after observation checks and numerical precision.

\paragraph{Room names require visual semantic grounding.}
Room options provide names, requiring models to infer room functions from layouts and furnishings. For GPT-6.1-sol and GPT-6-astra, room-function inference remains harder than local object recognition after N passes. This contrast highlights the need to connect visible objects with a room's overall layout and function.

\paragraph{Counting remains difficult after N and S pass.}
Counting errors persist for these models after both N and S pass. Completing observation checks does not ensure a consistent global count, highlighting cross-view instance identity and quantity integration as demands of spatial reasoning.

\begin{figure}[t]
  \centering
  \includegraphics[width=\linewidth]{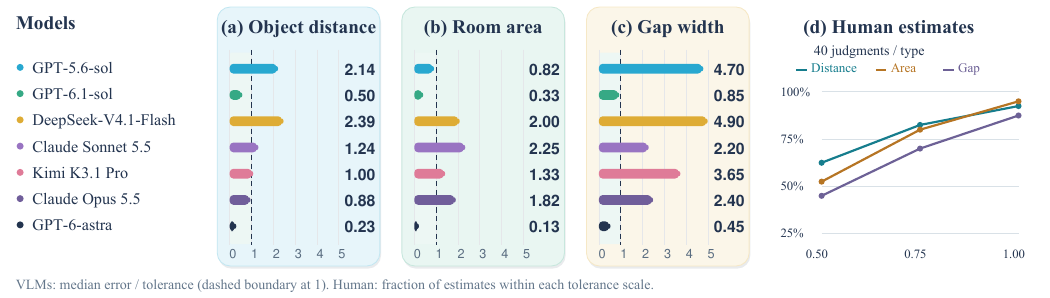}
  \caption{Spatial numerical precision. (a--c) Seven VLMs: median absolute error divided by tolerance among parseable answers. Values above one exceed the allowed error. (d) Human evaluators' estimates: percentages within each tolerance scale (40 judgments per type).}
  \label{fig:geometry}
\end{figure}

\paragraph{Free-space width is a precision bottleneck.}
Gap width has the most models with median error above tolerance (Fig.~\ref{fig:geometry}). Halving tolerances preserves the overall A/T ranking (Appendix~\ref{sec:tolerance_sensitivity}). Continuous errors reveal precision differences that binary scores can obscure.

\paragraph{Task-defined frames challenge direction judgments.}
Across all seven models, relative-direction errors exceed path-integration errors after N passes. These questions require constructing a frame from two reference objects to locate a third. Explicit representations of the frame's origin and forward axis may help these judgments.

\subsection{Object Relocation}
\label{sec:relocation_analysis}
\paragraph{Correct pickup is the largest relocation bottleneck.}
All seven models lose the most episodes before picking the correct object on the same relocation tasks. Later failures involve transferring that same object and verifying destination support (see the relocation milestones in Table~\ref{tab:relocation_milestones}). This shared pattern identifies target selection and correct pickup as the main relocation bottleneck.

\paragraph{Target selection and correction dominate early failures.}
Wrong-object pickup dominates early failures in every model. Feedback can also explain why a pick fails: the selected image point is on a floor or wall, or the arm cannot reach the object. Some models select another point or move the base but repeat the same error. These failures highlight the need to verify target identity and use feedback to correct subsequent actions.

\subsection{Oracle Input Diagnostics}
\label{sec:input_diagnostics}
We select 139 active-exploration tasks from G1--G4 in 26 scenes with recoverable final-answer inputs and target views. Four ground-truth (GT) settings test final judgments. \emph{GT retrieval} adds all task-relevant object frames from each model's own prior trajectory before answering. \emph{GT view supplementation} adds target views from other trajectories of the same task, including unseen targets. Both retain the original exploration context. \emph{GT trajectory RGB} replaces views and context with generated GT trajectory observations, retaining the question and public references. \emph{GT trajectory + sensors} adds radial depth, camera pose, and calibration jointly to the same RGB views and order. The baseline reuses original outcomes. Each GT setting produces one new answer without further actions, with construction-specific view counts and coverage. All seven models show similar trends across settings, so Table~\ref{tab:input_diagnostics} pools 973 evaluations per setting.

\begin{table}[t]
\centering
\small
\setlength{\tabcolsep}{4pt}
\caption{Correct answers across seven models on 139 questions.}
\label{tab:input_diagnostics}
\begin{tabular}{lr}
\toprule
Input & Correct answers \\
\midrule
Original episodes & 45.2\% (440/973) \\
GT retrieval & 49.4\% (481/973) \\
GT view supplementation & 58.4\% (568/973) \\
GT trajectory RGB & 64.9\% (631/973) \\
GT trajectory + sensors & 77.6\% (755/973) \\
\bottomrule
\end{tabular}
\end{table}

\paragraph{Additional target views improve final judgments.}
GT view supplementation improves final judgments more than GT retrieval in every model (Table~\ref{tab:input_diagnostics}). Earlier images cannot provide observations of targets missed during exploration. This pattern suggests that VLMs would benefit from recognizing which task-relevant targets remain unseen and seeking views that fill these gaps.

\paragraph{Complete views still require cross-view reasoning.}
GT trajectory RGB improves accuracy further with a jointly changed view set and exploration context. Remaining errors suggest that VLMs need to recognize the same object across viewpoints and combine separate observations into consistent judgments of object counts and spatial relations.

\paragraph{Explicit geometry helps connect views to physical space.}
With RGB views and order unchanged, adding depth, pose, and calibration jointly gives the highest accuracy. The gain highlights the value of relating objects across views to physical distances and camera positions, beyond recognizing them in individual images. Remaining errors suggest that VLMs also need to connect these measurements to the specific comparison or direction the question asks about.

\subsection{Human Analysis of Exploration Gaps}
\label{sec:human_validation}
Human evaluators categorize all 206 gap cases ($\ans=1$, $\nec=0$) across the 1,750 VLM episodes. Categories distinguish target observations, search scope, landmarks, and access or event verification. Of the 100 non-path cases, 85 miss required-object observations. One model correctly answers six mirrors while observing only five. Another observes all four side tables but leaves a required room unexplored. These cases and unseen-view gains (Section~\ref{sec:input_diagnostics}) motivate collecting target views and tracking instances separately from search coverage.

\section{Conclusion}
\bench{} evaluates exploration, spatial judgment, and goal-directed action through independent outcome and evidence checks. Human evaluators demonstrate successful task completion through the public observation and action interface. VLMs exhibit both correct outcomes with unmet task requirements and errors after stronger corroboration passes. Spatial-estimation and relocation failures further identify difficulties in grounding judgments and selecting the intended object. These findings motivate coherent cross-view instance representations, calibrated spatial judgment, and feedback-guided planning.

\FloatBarrier
\setlength{\bibsep}{3.5pt plus 1pt minus 0.5pt}
\bibliographystyle{ieeenat_fullname}
\bibliography{references}
\clearpage
\suppressfloats[t]
\beginappendix
\label{paper:appendix_start}
\setcounter{figure}{0}
\setcounter{table}{0}
\renewcommand{\thefigure}{S\arabic{figure}}
\renewcommand{\thetable}{S\arabic{table}}
\renewcommand{\theHfigure}{appendix.\arabic{figure}}
\renewcommand{\theHtable}{appendix.\arabic{table}}
\begingroup
\setlength{\tabcolsep}{3pt}
\section{Task Rules and Coverage}
\label{sec:appendix_tasks}
The tables document task groups, scene and task-type coverage, and the observations and actions required by each task. Outcome correctness (A), task-required evidence (N), and stronger corroboration (S) are assessed separately. S requires all N conditions plus task-specific corroboration, so S=1 implies N=1. Strict task success (T) requires A AND N.
\paragraph{Environment categories and scale.}
The 27 environments cover residential, office, retail and hospitality, community and leisure, and specialized workspaces. Each has 1--13 rooms (median 6) and 45--604 furnishings and other objects (median 159).

\begin{table}[t]
  \centering
  \small
  \caption{Task groups connect spatial questions to observable demands. Each task contributes to one primary group, although successful completion can require several demands together. Groups share requirements such as object search and spatial estimation.}
  \label{tab:capabilities}
  \begin{tabularx}{\textwidth}{L{0.22\textwidth}Yr}
    \toprule
    Diagnostic group & What the task asks the agent to establish & Tasks \\
    \midrule
    Semantic--spatial grounding & Identify surface or container contents, locate an object in a room, or compare rooms & 49 \\
    Spatial aggregation & Count objects over the declared scope or rooms containing a category & 38 \\
    Metric and comparative geometry & Estimate object distance, gap width, or room area, or compare which object is nearer & 77 \\
    Reference-frame integration & Answer in a specified frame or locate the episode start relative to the current pose & 56 \\
    Goal-directed spatial action & Find, pick, and place the designated instance at its destination & 30 \\
    \bottomrule
  \end{tabularx}
\end{table}

\begin{table}[t]
  \centering
  \small
  \caption{Distribution of the 250 tasks across 11 task types.}
  \label{tab:dataset}
  \begin{tabularx}{\columnwidth}{YrYr}
    \toprule
    Task type & Tasks & Task type & Tasks \\
    \midrule
    Surface contents & 20 & Room area & 20 \\
    Room localization & 29 & Gap width & 20 \\
    Object count & 20 & Relative direction & 26 \\
    Room count & 18 & Path integration & 30 \\
    Object distance & 20 & Object relocation & 30 \\
    Nearer-object comparison & 17 & & \\
    \bottomrule
  \end{tabularx}
\end{table}

\begin{table}[t]
  \centering
  \small
  \caption{N verifies task-required observations and interactions. S requires all N conditions and adds task-specific corroboration.}
  \label{tab:families}
  \begin{tabularx}{\textwidth}{L{0.18\textwidth}YY}
    \toprule
    Task group & What task-required evidence (N) checks & What stronger corroboration (S) adds \\
    \midrule
    G1: Grounding & The relevant target or contents are observed, with access checks; room comparisons also require exhaustive visual coverage and per-room counts & Qualifying target observations and room binding, with physical scope coverage for room comparisons \\
    G2: Aggregation & Required instances are observed over the task's declared search scope & Physical scope coverage and clear instance observations \\
    G3: Geometry & Required referents are observed, and room-area tasks include a target-room visit & Depth-based metric checks, uncertainty-aware nearest-object comparison, or room views from several directions \\
    G4: Reference frames & Required direction referents or path landmarks are observed & Reconstructed relative direction, or required room and viewpoint coverage for path integration \\
    G5: Spatial action & The designated source is successfully picked and then placed at the destination, with reveal checks when applicable & The released source has verified support from the designated destination \\
    \bottomrule
  \end{tabularx}
\end{table}

\begin{table}[t]
\centering\small
\setlength{\tabcolsep}{4pt}
\caption{Evaluation roles of task-required evidence conditions. One condition can support several roles.}
\label{tab:evidence_rule_roles}
\begin{tabularx}{\textwidth}{@{}L{0.20\textwidth}YY@{}}
\toprule
Task type & N condition & Evaluation role \\
\midrule
Surface contents & Observe the designated surface or container and listed objects & Relevant object information and local coverage \\
Room localization & Observe the target or accepted witness; satisfy access conditions and applicable room-comparison requirements & Object and room information, search coverage, and access \\
Object count & Observe counted instances and visually cover the search rooms & Instance information and search completeness \\
Room count & Observe instances, cover search rooms, and verify room assignment & Instance information and cross-room completeness \\
Object distance & Observe both referenced objects & Relevant objects for metric judgment \\
Nearer-object comparison & Observe the origin and all candidate objects & Relevant objects for comparison \\
Room area & Enter the target room & Target-space coverage and room entry \\
Gap width & Observe both furniture referents & Relevant objects for free-space judgment \\
Relative direction & Observe the origin, forward reference, and target & Object information for constructing the task frame \\
Path integration & Observe both specified landmarks & Completion of prescribed exploration \\
Object relocation & Pick and place the same accepted source at the destination; meet access conditions & Instance identity, interaction order, and access \\
\bottomrule
\end{tabularx}
\end{table}

Task-required evidence conditions (N) operationalize the observations and interactions required by each task instruction and are frozen before model evaluation. They serve related but distinct purposes: observing relevant objects, covering the declared search scope, and completing specified exploration or interaction. A condition can serve more than one purpose. In path integration, public self-motion information supports the directional answer, while landmark observations check completion of the prescribed exploration. Counting-scope coverage checks that the agent searches the required rooms, including places where further instances might occur.
\paragraph{Answer rules, viewpoint groups, and charged actions.}
The numeric scorer normalizes units. The reference distance $d$ is the nearest-integer horizontal straight-line distance between the two scene object centroids. Reference area $a$ is the room's recorded rectangular footprint (width times depth), rounded to the nearest integer and including occupied floor space. Answer tolerances are $\pm\max(2\,\mathrm{m},0.20d)$ and $\pm\max(5\,\mathrm{m}^2,0.25a)$. Gap width uses a tolerance of $\pm10\,\mathrm{cm}$. A records whether an estimate falls within the tolerance, while continuous errors describe its precision. Appendix~\ref{sec:tolerance_sensitivity} reports tighter-tolerance results, retaining the stored reference values. Geometric S checks retain their separate frozen task thresholds.

Viewpoints are grouped by room, position ($0.5$\,m), yaw ($30^\circ$), and camera horizon ($15^\circ$). Each group contributes once to coverage, so repeated observations from the same viewpoint do not inflate it.

Each task allows 25 high-level actions per physical room. Navigation, camera motion, and object interaction consume actions; retrieval and final answers do not. Internal walking and animation carry out each skill decision.
\paragraph{Visibility and instance identity.}
A qualifying observation passes the task's target-visibility rule. Regions belonging to the same instance are combined, including disconnected regions. The VLM receives RGB images 768 pixels wide and 432 pixels high. We inspect the delivered RGB views to select the visibility criterion before evaluation. The same criterion defines the scorer's clear-observation check. Repeated sightings are matched by instance ID across rounds. Visual room coverage counts camera entry or a qualifying observation of an object in that room. Physical room coverage requires a recorded camera pose inside the room.

\paragraph{Grounding and counting rules.}
For surface contents, N requires observing the designated surface or container and every listed object. S also checks that the objects have the required support relations and qualifying observations. For room localization, N requires observing the target or an accepted same-kind object in the answer room, with any required opening before a hidden target is revealed. S adds a qualifying target observation and a consistent object-to-room assignment. A verified copy made from the same object template can satisfy S through its own observation and room assignment. For object and room counts, N requires observing every counted instance and visually covering the specified search rooms. Room counts also check each instance's room assignment and the set of rooms containing targets. Room comparisons additionally check per-room counts. For these exhaustive-search tasks, S adds visits to every search room and qualifying observations of every counted instance.

\paragraph{Depth and geometric corroboration.}
Valid depth is finite and positive. The evaluator estimates each object's position from the qualifying view in which it occupies the largest image region and has a valid depth-derived point. It samples the region's center or a nearby point with representative depth. Distances are measured in the horizontal plane. Object-distance N observes both referents; S matches the reconstructed distance within the stored distance tolerance. Nearer-object N observes the origin and every candidate. S identifies the correct nearest candidate and separates its uncertainty interval from the runner-up. Object-position uncertainty is $\max(0.15,2/\sqrt{p})$ metres for mask area $p$; each distance sums the origin and candidate uncertainties. The nearest upper bound must be strictly below the runner-up lower bound.

Room-area N requires entering the target room. S requires at least three 90° yaw sectors observed inside it. The evaluator samples depth on a regular grid, with valid fraction at least 0.5 and an available median. Gap-width N observes both furniture referents. S uses finite depth-derived silhouette-edge points for both objects in the same observation, taking the minimum horizontal edge separation over eligible observations and comparing with the stored centimetre tolerance. Relative-direction N observes origin, forward reference, and target. S reconstructs the two horizontal bearings and matches the task sector within 22.5°. The object-based geometric checks also require qualifying object observations.

\paragraph{Landmark and relocation rules.}
For path integration, N checks whether both specified landmarks are observed, regardless of order. S adds the required room and distinct-viewpoint coverage. For relocation, N requires successfully picking an accepted source and subsequently placing that same object at the destination. Any required opening precedes target visibility or pickup. When several sources are acceptable, each source is checked through its complete pick-and-place sequence.

Relocation S requires all N conditions and a post-placement support readback matching the designated destination. The benchmark records this verified destination-support relation after release.

\paragraph{Public interaction and history.}
Each observation round supplies the inputs for the next skill or final answer. The interaction interface described in the main paper presents the task, sensor readings, action feedback, reference images, and the most recent 20 visual rounds. All image sources share a limit of 50 images per decision. Retrieved camera images and their original text can supplement this history. Intermediate camera frames from the preceding action are shown only at the first decision in a round. A path visualization and the latest camera image follow. Image-point actions refer only to that latest image.

At the first decision, the agent may retain selected intermediate frames. The default retains none, and the selection remains unchanged for the round. Before executing a skill or submitting an answer or completion report, the agent may retrieve earlier observations at most twice, with at most 12 archived images in total. Empty or invalid retrieval requests count toward this limit. Retrieval updates the context without changing the environment or advancing the observation round. Navigation and interaction skills produce the next round's inputs. An answer or completion report ends the episode and triggers independent evaluation.

Latest camera images and retained intermediate frames enter the recent visual history and retrieval archive. Unselected frames remain available for offline inspection after their first delivery. Dialogue history and the latest path visualization remain in the input. Task ground truth and scores are withheld from the agent.

Skills also check body collisions and target reachability, with a nominal horizontal manipulation reach of 1.5\,m. Appendix~\ref{sec:appendix_configuration} collects the complete input limits, body dimensions, collision margins, reach tolerance, and inference settings.
\begin{figure}[t]
  \centering
  \paperfigure{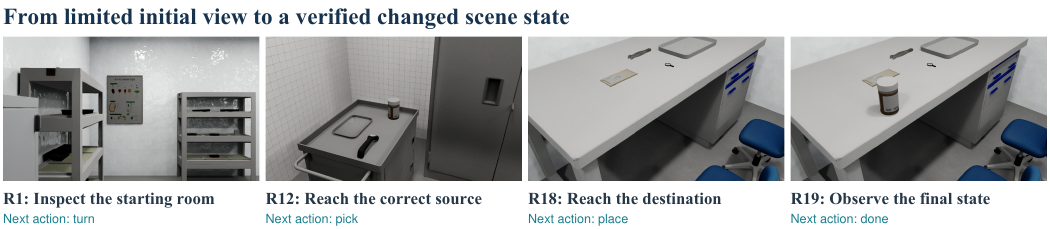}
  \caption{Saved observations from a successful GPT-6-astra relocation episode (scene\_148). Rounds 1, 12, 18, and 19 show the initial view, pre-pick view, pre-place view, and post-place view before completion. The episode contains 18 charged actions and 19 rounds.}
  \label{fig:real_episode}
\end{figure}
The accompanying code and task submission includes a complete anonymized record of the episode in Figure~\ref{fig:real_episode}, with observations, actions, feedback, and evaluation checks.
\section{Measurement-Level Positioning}
\label{sec:appendix_positioning}
The main paper groups related work by the capabilities it evaluates. Table~\ref{tab:positioning} compares what these evaluations measure and how they relate to our episode-level checks.
\begin{table}[t]
  \centering
  \small
  \caption{Measurement scope of related work and ROAM-Bench. Our setting connects active observation, spatial judgment, action, and task-required evidence within an episode.}
  \label{tab:positioning}
  \begin{tabularx}{\textwidth}{L{0.23\textwidth}YY}
    \toprule
    Work & Main measurement object & Connection to our setting \\
    \midrule
    EmbodiedQA~\cite{das2018embodied} / OpenEQA~\cite{majumdar2024openeqa} & Answers from first-person observations or exploration & Shared question-conditioned information gathering \\
    EXPRESS-Bench~\cite{jiang2025express} & VLM-scored correctness and grounding in the final observation & Final-image assessment versus trajectory-level task requirements and corroboration \\
    ESI-Bench~\cite{hong2026esibench} & Spatial QA through perception, locomotion, and manipulation & Broad spatial faculties versus task-required evidence, geometric checks, and ordered relocation with destination-support checks \\
    E3VS-Bench~\cite{sakamoto2026e3vs} & Viewpoint-dependent evidence acquisition & Shared active search. We also check spatial outcomes and task-required evidence \\
    VSI-Bench~\cite{yang2024thinking} / MindCube~\cite{wang2025mindcube} & Spatial inference from supplied video or limited views & Reasoning given views versus choosing observations during a task \\
    MMSI-Bench~\cite{yang2025mmsi} & Multi-image answers, reference-assisted reasoning diagnosis, and downstream correlations & Supplied-image diagnosis versus acquired observations and actions within one episode \\
    Room-to-Room~\cite{anderson2017vln} / Beyond Spatial Benchmarks~\cite{huang2026spatialnav} & Navigation success and spatial-reasoning transfer & Route/transfer outcomes versus task-required evidence in an episode \\
    BEHAVIOR-Spatial~\cite{behavior2026spatial} & Relationships among spatial QA, grounding, and robot action & Across-evaluation relationships versus same-episode outcome and evidence checks \\
    EmbodiedBench~\cite{yang2025embodiedbench} / Embodied Agent Arena~\cite{huang2026arena} & Broad embodied task success and capability diagnostics & Shared broad task coverage; our spatial tasks connect outcomes with episode-level observation and interaction checks \\
    \bench{} & Spatial outcomes with required observations and interactions & Outcome correctness, task-required evidence, and stronger corroboration are checked separately \\
    \bottomrule
  \end{tabularx}
\end{table}

\section{Results by Task Type}
\label{sec:operation_profiles}
Table~\ref{tab:operation_profiles} reports A and T for each task type across all seven models. Each task contributes once to its primary group. Appendix~\ref{sec:appendix_bottlenecks} connects these results to spatial-answer errors and relocation stages.
\begin{table}[t]
\centering\small
\setlength{\tabcolsep}{3pt}
\caption{A / T percentages by task type across seven models. Each task contributes once to its primary group.}
\label{tab:operation_profiles}
\begin{tabularx}{\textwidth}{lYrrrrrrrr}
\toprule
Group & Task type & Tasks & \shortstack{GPT-5.6-\\sol} & \shortstack{GPT-6.1-\\sol} & \shortstack{DeepSeek\\V4.1-Flash} & \shortstack{Claude\\Sonnet 5.5} & \shortstack{Kimi\\K3.1 Pro} & \shortstack{Claude\\Opus 5.5} & \shortstack{GPT-6-\\astra} \\
\midrule
G1 & Surface contents & 20 & 60/60 & 90/90 & 10/10 & 75/75 & 85/85 & 95/95 & 100/100 \\
G1 & Room localization & 29 & 34.5/34.5 & 37.9/27.6 & 31/27.6 & 65.5/55.2 & 41.4/37.9 & 58.6/44.8 & 55.2/48.3 \\
G2 & Object count & 20 & 15/10 & 35/30 & 20/15 & 30/20 & 35/25 & 40/40 & 50/50 \\
G2 & Room count & 18 & 22.2/16.7 & 50/38.9 & 16.7/16.7 & 38.9/33.3 & 27.8/16.7 & 50/38.9 & 66.7/55.6 \\
G3 & Object distance & 20 & 25/10 & 70/60 & 25/15 & 40/35 & 55/35 & 70/65 & 85/85 \\
G3 & Nearer-object comparison & 17 & 35.3/0 & 88.2/76.5 & 17.6/0 & 58.8/35.3 & 52.9/17.6 & 70.6/41.2 & 100/94.1 \\
G3 & Room area & 20 & 55/45 & 90/85 & 15/15 & 20/10 & 35/35 & 30/25 & 100/100 \\
G3 & Gap width & 20 & 10/10 & 60/60 & 5/5 & 10/10 & 15/15 & 10/10 & 75/75 \\
G4 & Relative direction & 26 & 19.2/3.8 & 42.3/23.1 & 11.5/0 & 19.2/7.7 & 30.8/19.2 & 38.5/26.9 & 61.5/61.5 \\
G4 & Path integration & 30 & 86.7/16.7 & 96.7/50 & 46.7/6.7 & 83.3/33.3 & 90/20 & 96.7/33.3 & 100/86.7 \\
G5 & Object relocation & 30 & 6.7/6.7 & 10/10 & 3.3/3.3 & 30/30 & 13.3/13.3 & 33.3/30 & 33.3/30 \\
\bottomrule
\end{tabularx}
\end{table}

\section{Task Difficulty Across Models}
\label{sec:complete250_difficulty}
We measure each task's answer accuracy across the seven models, with each model contributing once. The denominator is seven for every task.
\begin{table}[t]
\centering\small
\caption{Fractions of per-task answer-accuracy variance associated with task type, scene, search and object requirements, and action budget across seven models. Every task has denominator seven. These in-sample associations can overlap.}
\label{tab:complete250_difficulty}
\begin{tabular}{lr}
\toprule
Grouping feature & $\eta^2$ \\
\midrule
Task type & 0.466 \\
Scene & 0.197 \\
Scene room count & 0.053 \\
Required visited rooms & 0.009 \\
Required search rooms & 0.116 \\
Required object instances & 0.022 \\
Referenced instances & 0.067 \\
Action budget & 0.053 \\
\bottomrule
\end{tabular}
\end{table}

Task type and scene are associated with 0.466 and 0.197 of the observed between-task accuracy variance ($\eta^2$), respectively. Their additive fit has in-sample $R^2=0.545$. The associations overlap because tasks share both type and scene.

We hold out each of the 27 scenes in turn and estimate task-type means from the remaining scenes. These means reduce squared prediction error by 41.46\% relative to the training-mean baseline, showing that task type helps predict difficulty across the benchmark environments.

Twenty-nine tasks receive no correct answer from any of the seven models. These shared failures indicate task difficulties that persist across models with different overall scores. Appendix~\ref{sec:appendix_visual} provides human-verified analysis of object visibility and behavior across all seven models.

\paragraph{Overall performance comparison.}
GPT-6-astra achieves A=73.20\% and T=69.20\%, with 43.62 actions per episode, including failures. Its paired T advantage over GPT-6.1-sol is 22.40 percentage points (10,000 whole-scene resamples, percentile range 15.83--28.83).

\section{Episode-Level Diagnostics}
\label{sec:posthoc_diagnostics}

The VLM analyses in Appendices~\ref{sec:posthoc_diagnostics}--\ref{sec:appendix_bottlenecks} use the same 1,750 episodes (250 tasks per model). Here we summarize outcome gaps, relocation milestones, numerical errors, and action costs. Relocation uses the 210-episode subset (30 tasks per model). Human review categorizes all 206 gap cases (A=1,N=0) here, examines localizability and behavior in 307 sampled episodes (Appendix~\ref{sec:visual_review_protocol}), and directly audits N/S in 200 episodes (Appendix~\ref{sec:ns_human_audit}). Appendix~\ref{sec:appendix_bottlenecks} breaks down the scores and relocation stages by model.

Ranges are the 2.5th and 97.5th percentiles of 10,000 whole-scene bootstrap resamples, keeping tasks and model observations from the same environment together.

\subsection{Correctness and Task Completion}
\label{sec:appendix_gap_attribution}
Excluding path integration, 100 of 1,540 episodes yield correct outcomes with unmet task requirements (6.5 percentage points). In model order from Table~\ref{tab:support_gap}, these counts are 17, 16, 10, 16, 18, 17, and 6, each over 220 tasks. They concern required target observations, search coverage, or access and interaction checks. Path integration adds 106 cases of unmet landmark exploration, while start and final poses supply direction information. Table~\ref{tab:support_gap} reports per-model gap cases (A=1,N=0), their fraction of all episodes (G), and their fraction of correct outcomes.

\begin{table}[t]
\centering\small
\caption{Gap cases are correct outcomes with unmet task-required evidence (A=1,N=0). G divides this count by all 250 episodes per model; gap among correct divides it by A-passing episodes. Brackets show whole-scene bootstrap percentiles.}
\label{tab:support_gap}
\begin{tabularx}{\textwidth}{Yrrll}
\toprule
Model & Correct (n) & Gap cases (n) & G (\%) [scene percentiles] & Gap among correct (\%) \\
\midrule
GPT-5.6-sol & 86 & 38 & 15.20 [8.66, 21.96] & 44.19 [28.84, 57.83] \\
GPT-6.1-sol & 147 & 30 & 12.00 [7.00, 17.23] & 20.41 [12.61, 28.05] \\
DeepSeek-V4.1-Flash & 48 & 22 & 8.80 [4.22, 13.83] & 45.83 [24.49, 63.93] \\
Claude Sonnet 5.5 & 110 & 31 & 12.40 [6.71, 19.52] & 28.18 [16.53, 40.77] \\
Kimi K3.1 Pro & 110 & 39 & 15.60 [8.98, 22.46] & 35.45 [21.65, 48.54] \\
Claude Opus 5.5 & 136 & 36 & 14.40 [8.23, 20.82] & 26.47 [15.79, 36.52] \\
GPT-6-astra & 183 & 10 & 4.00 [1.15, 7.47] & 5.46 [1.57, 10.19] \\
\bottomrule
\end{tabularx}
\end{table}

\paragraph{Required-object observations account for most non-landmark gaps.}
We manually inspect and categorize all 206 gap cases (A=1,N=0) using their task requirements and corresponding observation and interaction records. Tables~\ref{tab:supp-gap-reasons-task} and~\ref{tab:supp-gap-reasons-model} separate object observations, search scope, landmarks, and access or event verification. One episode belongs to two categories. Of the 100 non-path cases, 85 fail required-object observation checks. Target coverage is thus the principal non-landmark shortfall, distinct from search completeness and prescribed procedures.

\begin{table}[t]
\centering
\small
\setlength{\tabcolsep}{1.5pt}
\caption{Manual categorization of all 206 gap cases ($A=1,N=0$). Obj.: incomplete relevant-object observations. Scope: unmet prescribed room coverage. Land.: unobserved required landmarks. Inter.: unmet or unverified access/event checks. These seven episodes have an unverified initial placement relation; one also lacks the required ordered events. Columns overlap. Multi. counts episodes in at least two categories.}
\label{tab:supp-gap-reasons-task}
\begin{tabular}{@{}lrrrrrr@{}}
\toprule
Task type & \shortstack{Gap\\cases} & Obj. & Scope & Land. & Inter. & Multi. \\
\midrule
Room localization & 14 & 9 & 0 & 0 & 5 & 0 \\
Object count & 7 & 5 & 3 & 0 & 0 & 1 \\
Room count & 10 & 10 & 0 & 0 & 0 & 0 \\
Object distance & 13 & 13 & 0 & 0 & 0 & 0 \\
Nearer-object comparison & 27 & 27 & 0 & 0 & 0 & 0 \\
Room area & 6 & 0 & 6 & 0 & 0 & 0 \\
Relative direction & 21 & 21 & 0 & 0 & 0 & 0 \\
Path integration & 106 & 0 & 0 & 106 & 0 & 0 \\
Object relocation & 2 & 0 & 0 & 0 & 2 & 0 \\
\midrule
All & 206 & 85 & 9 & 106 & 7 & 1 \\
\bottomrule
\end{tabular}
\end{table}

\begin{table}[t]
\centering
\small
\setlength{\tabcolsep}{2.5pt}
\caption{Manual categorization by model of gap cases ($A=1,N=0$). Categories and overlap follow Table~\ref{tab:supp-gap-reasons-task}.}
\label{tab:supp-gap-reasons-model}
\begin{tabular}{lrrrrrr}
\toprule
Model & \shortstack{Gap\\cases} & Obj. & Scope & Land. & Inter. & Multi. \\
\midrule
GPT-5.6-sol & 38 & 15 & 2 & 21 & 0 & 0 \\
GPT-6.1-sol & 30 & 14 & 1 & 14 & 1 & 0 \\
DeepSeek-V4.1-Flash & 22 & 10 & 0 & 12 & 0 & 0 \\
Claude Sonnet 5.5 & 31 & 12 & 4 & 15 & 1 & 1 \\
Kimi K3.1 Pro & 39 & 16 & 1 & 21 & 1 & 0 \\
Claude Opus 5.5 & 36 & 14 & 1 & 19 & 2 & 0 \\
GPT-6-astra & 10 & 4 & 0 & 4 & 2 & 0 \\
\midrule
All & 206 & 85 & 9 & 106 & 7 & 1 \\
\bottomrule
\end{tabular}
\end{table}

\paragraph{Correct counts with incomplete exploration.}
Two recorded examples distinguish target observations from search coverage. In scene\_107, GPT-6.1-sol reports the correct total of six mirrors, but its delivered observations cover only five required mirror instances, omitting the guest-lounge mirror. In scene\_139, Kimi K3.1 Pro reports the correct total of four side tables and observes all four, but does not complete the required room coverage, missing the steam room. Both pass A and fail N and T. The first lacks a required target observation, while the second fails a prescribed search condition despite observing all counted targets.

\subsection{Relocation Milestones}
\label{sec:appendix_relocation_milestones}

\begin{figure}[t]
  \centering
  \includegraphics[width=\linewidth]{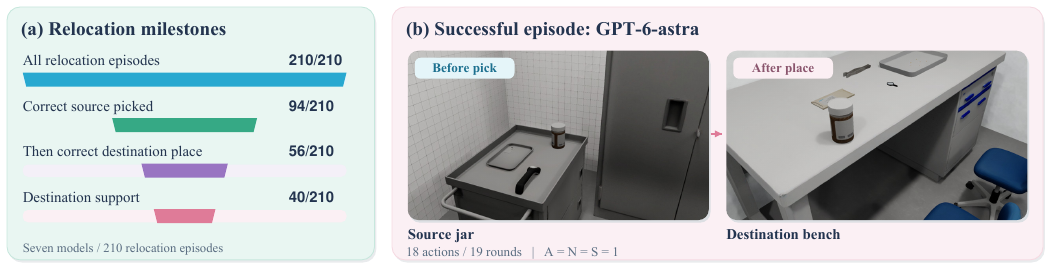}
  \caption{Relocation across seven models. Left: successive stages over 210 episodes. Placement requires a preceding pick of the same acceptable object. Destination support adds a post-placement check of the released object. Right: pre-pick and post-place views from a successful GPT-6-astra episode.}
  \label{fig:relocation_funnel}
\end{figure}

Across the 210 relocation episodes, 94 successfully pick an acceptable target, 56 subsequently place the same object at the specified destination, and 40 pass the destination-support check after placement. Milestones use recorded event order, including events within one observation round, following the relocation rules in Appendix~\ref{sec:appendix_tasks}.

\begin{table}[t]
\centering\small
\setlength{\tabcolsep}{3pt}
\caption{Relocation milestones across 30 tasks per model, following Appendix~\ref{sec:appendix_tasks}. Pick and Place count interaction milestones; Support adds the post-placement destination-support check. A and T score outcome correctness and strict task success.}
\label{tab:relocation_milestones}
\begin{tabularx}{\columnwidth}{Yrrrr}
\toprule
Model & Pick & Place & Support & A / T \\
\midrule
GPT-5.6-sol & 9 & 3 & 2 & 2 / 2 \\
GPT-6.1-sol & 15 & 6 & 3 & 3 / 3 \\
DeepSeek-V4.1-Flash & 5 & 1 & 1 & 1 / 1 \\
Claude Sonnet 5.5 & 18 & 13 & 9 & 9 / 9 \\
Kimi K3.1 Pro & 14 & 6 & 4 & 4 / 4 \\
Claude Opus 5.5 & 17 & 12 & 10 & 10 / 9 \\
GPT-6-astra & 16 & 15 & 11 & 10 / 9 \\
\bottomrule
\end{tabularx}
\end{table}

These milestones count pickup, placement, and support events. N and S additionally require the applicable access checks, while A checks the final goal state. Later actions can change the final state after a successful placement.

\paragraph{Wrong-object pickup and recovery.}
We inspect all 210 episodes using the task's acceptable target instances and the recorded pick events. A wrong-object pickup is a successful pick of an object outside this acceptable set. Of the 116 episodes without any correct pickup, 102 contain a wrong-object pickup without later correction (87.9\%). Thus, target selection errors that remain uncorrected account for most failures before correct pickup. Per-model counts are also reported in Table~\ref{tab:bottleneck_execution}.

Among the 116 episodes without a correct pickup, 36 contain a failed pick action. Fourteen contain failed picks at two or more distinct action steps. These categories can overlap with wrong-object pickup. Across all 210 episodes, 20 reach a correct pickup after a failed pick action, and 15 reach a correct pickup after a wrong-object pickup. These events show that recovery is possible, while the 102 uncorrected wrong-object episodes identify target verification and correction as major weaknesses.

\paragraph{Responses to explicit failure feedback.}
The recorded actions illustrate how a strategy change can still fail to address the problem. In a DeepSeek-V4.1-Flash proof-card episode, two successive pick actions select different image points. Both receive feedback that the point lies on a floor, wall, or ceiling and the arm does not move. In a GPT-5.6-sol kettle episode, the model moves sideways and forward after reachability failures. The next pick still reports no reachable arm pose, and a later successful pick holds a different object. These episodes show attempts to adjust actions, but inadequate correction of the selected point or target identity. They motivate using feedback to decide whether to change the grasp point, approach position, or selected object, then checking the result in the next image.

\subsection{Numerical Estimation Errors}
\label{sec:appendix_geometry}

Answer accuracy alone does not show the size of numerical errors. Each model answers the same 20 tasks in each of three numerical types. We convert parsed answers to the task's scoring units and measure absolute error, relative error, and absolute error divided by the answer tolerance. A normalized error of one is the acceptance boundary. Error summaries use parseable answers, with coverage reported for every model and task type.

\begin{table}[t]

\centering\footnotesize

\caption{Numerical estimates evaluated with the task-defined acceptance rule. Parsing coverage accompanies every conditional error summary.}
\begin{tabularx}{\textwidth}{Ylrllr}

\toprule

Task & Model & Parsed / all & \shortstack{Mean absolute error\\scene percentiles} & \shortstack{Mean relative error (\%)\\scene percentiles} & \shortstack{Median absolute error\\/ tolerance} \\

\midrule

Object distance (m) & GPT-5.6-sol & 20 / 20 & 5.15 [3.63, 7.10] & 52.55 [42.67, 61.14] & 2.136 \\

Object distance (m) & GPT-6.1-sol & 19 / 20 & 1.56 [0.71, 2.45] & 17.05 [7.96, 26.41] & 0.500 \\

Object distance (m) & DeepSeek-V4.1-Flash & 20 / 20 & 5.72 [4.06, 7.84] & 58.50 [48.96, 68.12] & 2.386 \\

Object distance (m) & Claude Sonnet 5.5 & 20 / 20 & 2.91 [1.85, 3.92] & 28.58 [19.18, 37.21] & 1.236 \\

Object distance (m) & Kimi K3.1 Pro & 19 / 20 & 3.33 [2.27, 5.13] & 32.38 [27.60, 39.97] & 1.000 \\

Object distance (m) & Claude Opus 5.5 & 20 / 20 & 2.27 [1.52, 3.20] & 24.11 [17.42, 30.35] & 0.875 \\

Object distance (m) & GPT-6-astra & 20 / 20 & 0.86 [0.40, 1.60] & 9.18 [4.46, 15.87] & 0.225 \\

Room area (m$^2$) & GPT-5.6-sol & 20 / 20 & 13.60 [7.40, 20.23] & 30.21 [18.53, 40.83] & 0.825 \\

Room area (m$^2$) & GPT-6.1-sol & 20 / 20 & 4.09 [1.97, 6.24] & 10.47 [5.66, 14.88] & 0.328 \\

Room area (m$^2$) & DeepSeek-V4.1-Flash & 19 / 20 & 25.32 [15.00, 38.41] & 51.06 [35.85, 69.13] & 2.000 \\

Room area (m$^2$) & Claude Sonnet 5.5 & 20 / 20 & 27.73 [18.56, 38.19] & 53.83 [45.95, 61.10] & 2.254 \\

Room area (m$^2$) & Kimi K3.1 Pro & 20 / 20 & 19.30 [9.69, 30.88] & 35.24 [24.48, 46.27] & 1.333 \\

Room area (m$^2$) & Claude Opus 5.5 & 20 / 20 & 22.45 [12.61, 34.84] & 40.80 [29.88, 51.89] & 1.820 \\

Room area (m$^2$) & GPT-6-astra & 20 / 20 & 1.45 [0.99, 2.05] & 4.39 [2.78, 6.73] & 0.133 \\

Gap width (cm) & GPT-5.6-sol & 20 / 20 & 49.35 [33.33, 75.92] & 60.03 [39.48, 94.32] & 4.700 \\

Gap width (cm) & GPT-6.1-sol & 20 / 20 & 12.05 [7.81, 17.62] & 13.68 [9.00, 18.61] & 0.850 \\

Gap width (cm) & DeepSeek-V4.1-Flash & 18 / 20 & 64.33 [38.04, 107.93] & 86.57 [46.79, 160.72] & 4.900 \\

Gap width (cm) & Claude Sonnet 5.5 & 20 / 20 & 24.90 [19.82, 33.27] & 29.57 [23.24, 39.23] & 2.200 \\

Gap width (cm) & Kimi K3.1 Pro & 18 / 20 & 32.78 [25.15, 44.58] & 41.22 [30.83, 56.14] & 3.650 \\

Gap width (cm) & Claude Opus 5.5 & 20 / 20 & 25.60 [20.76, 33.25] & 30.63 [24.19, 41.42] & 2.400 \\

Gap width (cm) & GPT-6-astra & 20 / 20 & 9.25 [4.84, 17.67] & 10.06 [5.87, 18.03] & 0.450 \\

\bottomrule

\end{tabularx}

\end{table}

Object distance, room area, and gap width are reported in metres, square metres, and centimetres, respectively. Normalizing by tolerance relates these different quantities to their acceptance rules. Gap width has the most models with median errors above tolerance, identifying physical free-space estimation as a priority for geometric reasoning.

\paragraph{Human numerical estimates.}
Two human evaluators independently estimate all 20 tasks in each numerical type, giving 40 judgments per type and 120 overall. They receive the public task materials and standardized RGB views containing the targets, with depth, reference values, and tolerances withheld. This experiment evaluates numerical estimation from supplied views, separately from the 100-task human active-exploration evaluation.

Table~\ref{tab:human_tolerance} reports absolute and relative errors, acceptance rates, and agreement. Default tolerances accept 110/120 estimates (91.7\%), compared with 93/120 (77.5\%) and 64/120 (53.3\%) at the tighter scales. Acceptance spans 87.5--95.0\% across the three task types, with ICC(2,1) of 0.76--0.84. These results place the binary thresholds against human visual estimates. The continuous errors and tighter-score results preserve a separate measure of precision.

\begin{table}[t]
\centering\footnotesize
\setlength{\tabcolsep}{3.5pt}
\caption{Independent human numerical estimates under the three tolerance scales. Two human evaluators each judge 20 tasks per type. MAE uses task units. ICC(2,1) measures single-rater absolute agreement under a two-way random-effects model. The pooled row combines judgments across types.}
\label{tab:human_tolerance}
\begin{tabular}{@{}lccrrrrr@{}}
\toprule
Task & Tasks / judgments & MAE & Median relative (\%) & $1.00$ pass (\%) & $0.75$ pass (\%) & $0.50$ pass (\%) & ICC(2,1) \\
\midrule
Object distance & 20 / 40 & 1.07 m & 9.4 & 37/40 (92.5) & 33/40 (82.5) & 25/40 (62.5) & 0.84 \\
Room area & 20 / 40 & 3.62 m$^2$ & 10.7 & 38/40 (95.0) & 32/40 (80.0) & 21/40 (52.5) & 0.79 \\
Gap width & 20 / 40 & 8.34 cm & 11.5 & 35/40 (87.5) & 28/40 (70.0) & 18/40 (45.0) & 0.76 \\
Overall & 60 / 120 & -- & 10.6 & 110/120 (91.7) & 93/120 (77.5) & 64/120 (53.3) & -- \\
\bottomrule
\end{tabular}
\end{table}

\paragraph{Sensitivity to answer tolerances.}
\label{sec:tolerance_sensitivity}
We rescore the seven models' 1,750 recorded episodes with distance, room-area, and gap-width answer tolerances scaled by 1.00, 0.75, and 0.50. Answers, parsed values, trajectories, N/S checks, and other task rules remain fixed. The original scale reproduces all main-result counts. Unparseable answers remain incorrect. Scores and paired differences use the 2.5th and 97.5th percentiles of 10,000 whole-scene bootstrap resamples across the same 27 scenes (seed 20261113). Each resampled scene retains all of its tasks and models.

Table~\ref{tab:tolerance_sensitivity} shows that halving all three tolerances lowers overall T by 0.8--4.0 percentage points. Overall A and T rankings retain their order at both tighter settings, with GPT-6-astra highest and the Sonnet--Kimi A tie broken in favor of Sonnet. Tables~\ref{tab:tolerance_by_task}--\ref{tab:tolerance_gap} give each numerical task type. Distance and area have pairwise rank reversals, reported with paired intervals in Table~\ref{tab:tolerance_rank_changes}. Gap width preserves every originally strict ordering, with some ties separating at the half scale. Overall rankings are stable, while task-level errors reveal finer differences in spatial precision.

\begin{table}[t]
\centering\small
\setlength{\tabcolsep}{5pt}
\caption{Overall tolerance sensitivity on 250 tasks per model. Each cell gives A (top) and T (bottom), as percentages with 95\% whole-scene bootstrap intervals. Paired T drops are percentage points. Only the distance, area, and gap answer tolerances change.}
\label{tab:tolerance_sensitivity}
\begin{tabular}{@{}lcccc@{}}
\toprule
Model & $1.00$ A / T & $0.75$ A / T & $0.50$ A / T & $0.50$ T drop \\
\midrule
GPT-5.6-sol & \shortstack{34.4 [26.5, 42.8]\\19.2 [13.4, 25.2]} & \shortstack{32.4 [23.8, 41.5]\\17.6 [11.8, 23.5]} & \shortstack{31.2 [23.1, 39.9]\\16.4 [11.0, 22.0]} & 2.8 [0.9, 4.8] \\
GPT-6.1-sol & \shortstack{58.8 [50.8, 66.5]\\46.8 [40.0, 53.6]} & \shortstack{56.8 [48.4, 64.7]\\45.6 [38.7, 52.4]} & \shortstack{53.6 [45.4, 61.4]\\42.8 [35.7, 50.0]} & 4.0 [1.7, 6.5] \\
DeepSeek-V4.1-Flash & \shortstack{19.2 [14.6, 24.4]\\10.4 [6.6, 14.8]} & \shortstack{18.0 [13.4, 23.3]\\9.6 [5.6, 14.2]} & \shortstack{18.0 [13.4, 23.3]\\9.6 [5.6, 14.2]} & 0.8 [0.0, 2.0] \\
Claude Sonnet 5.5 & \shortstack{44.0 [36.9, 51.1]\\31.6 [25.5, 37.4]} & \shortstack{42.8 [36.3, 49.2]\\31.2 [24.8, 37.1]} & \shortstack{42.0 [35.7, 48.2]\\30.4 [24.3, 35.8]} & 1.2 [0.0, 2.6] \\
Kimi K3.1 Pro & \shortstack{44.0 [37.8, 49.8]\\28.4 [22.2, 35.1]} & \shortstack{42.4 [36.4, 48.2]\\27.6 [21.3, 34.4]} & \shortstack{41.2 [35.3, 46.9]\\26.8 [20.8, 33.3]} & 1.6 [0.4, 3.1] \\
Claude Opus 5.5 & \shortstack{54.4 [47.9, 60.6]\\40.0 [33.9, 46.4]} & \shortstack{52.0 [44.9, 58.6]\\38.0 [32.1, 44.0]} & \shortstack{50.0 [43.1, 56.7]\\36.0 [30.7, 41.5]} & 4.0 [1.3, 7.7] \\
GPT-6-astra & \shortstack{73.2 [65.8, 80.4]\\69.2 [61.1, 77.0]} & \shortstack{72.8 [65.5, 79.8]\\68.8 [60.8, 76.4]} & \shortstack{71.2 [63.9, 78.2]\\67.2 [59.3, 74.7]} & 2.0 [0.5, 3.7] \\
\bottomrule
\end{tabular}
\end{table}

\begin{table}[t]
\centering\small
\setlength{\tabcolsep}{5pt}
\caption{Object distance tolerance sensitivity on 20 tasks per model. Each cell gives A (top) and T (bottom), as percentages with 95\% whole-scene bootstrap intervals.}
\label{tab:tolerance_by_task}
\begin{tabular}{@{}lccc@{}}
\toprule
Model & $1.00$ A / T & $0.75$ A / T & $0.50$ A / T \\
\midrule
GPT-5.6-sol & \shortstack{25.0 [5.6, 50.0]\\10.0 [0.0, 26.7]} & \shortstack{15.0 [0.0, 33.3]\\5.0 [0.0, 13.3]} & \shortstack{15.0 [0.0, 33.3]\\5.0 [0.0, 13.3]} \\
GPT-6.1-sol & \shortstack{70.0 [45.0, 94.1]\\60.0 [30.8, 84.6]} & \shortstack{65.0 [37.5, 89.2]\\60.0 [30.8, 84.6]} & \shortstack{50.0 [23.1, 78.3]\\50.0 [23.1, 78.3]} \\
DeepSeek-V4.1-Flash & \shortstack{25.0 [6.7, 42.9]\\15.0 [0.0, 27.3]} & \shortstack{10.0 [0.0, 26.7]\\5.0 [0.0, 13.3]} & \shortstack{10.0 [0.0, 26.7]\\5.0 [0.0, 13.3]} \\
Claude Sonnet 5.5 & \shortstack{40.0 [16.7, 62.5]\\35.0 [11.8, 56.5]} & \shortstack{40.0 [16.7, 62.5]\\35.0 [11.8, 56.5]} & \shortstack{35.0 [16.7, 54.2]\\30.0 [11.1, 47.1]} \\
Kimi K3.1 Pro & \shortstack{55.0 [36.4, 71.4]\\35.0 [11.1, 56.2]} & \shortstack{40.0 [14.3, 58.8]\\30.0 [7.1, 50.0]} & \shortstack{25.0 [0.0, 40.9]\\20.0 [0.0, 35.7]} \\
Claude Opus 5.5 & \shortstack{70.0 [46.7, 93.8]\\65.0 [40.0, 88.9]} & \shortstack{45.0 [25.0, 63.6]\\45.0 [25.0, 63.6]} & \shortstack{30.0 [15.0, 50.0]\\30.0 [15.0, 50.0]} \\
GPT-6-astra & \shortstack{85.0 [63.6, 100.0]\\85.0 [63.6, 100.0]} & \shortstack{85.0 [63.6, 100.0]\\85.0 [63.6, 100.0]} & \shortstack{85.0 [63.6, 100.0]\\85.0 [63.6, 100.0]} \\
\bottomrule
\end{tabular}
\end{table}

\begin{table}[t]
\centering\small
\setlength{\tabcolsep}{5pt}
\caption{Room area tolerance sensitivity on 20 tasks per model. Each cell gives A (top) and T (bottom), as percentages with 95\% whole-scene bootstrap intervals.}
\label{tab:tolerance_area}
\begin{tabular}{@{}lccc@{}}
\toprule
Model & $1.00$ A / T & $0.75$ A / T & $0.50$ A / T \\
\midrule
GPT-5.6-sol & \shortstack{55.0 [33.3, 77.3]\\45.0 [23.1, 66.7]} & \shortstack{40.0 [17.4, 64.7]\\30.0 [8.3, 53.3]} & \shortstack{25.0 [6.9, 50.0]\\15.0 [0.0, 35.0]} \\
GPT-6.1-sol & \shortstack{90.0 [76.0, 100.0]\\85.0 [70.4, 100.0]} & \shortstack{80.0 [57.9, 100.0]\\80.0 [57.9, 100.0]} & \shortstack{60.0 [38.5, 83.3]\\60.0 [38.5, 83.3]} \\
DeepSeek-V4.1-Flash & \shortstack{15.0 [0.0, 31.6]\\15.0 [0.0, 31.6]} & \shortstack{15.0 [0.0, 31.6]\\15.0 [0.0, 31.6]} & \shortstack{15.0 [0.0, 31.6]\\15.0 [0.0, 31.6]} \\
Claude Sonnet 5.5 & \shortstack{20.0 [5.0, 36.8]\\10.0 [0.0, 26.7]} & \shortstack{5.0 [0.0, 17.6]\\5.0 [0.0, 17.6]} & \shortstack{0.0 [0.0, 0.0]\\0.0 [0.0, 0.0]} \\
Kimi K3.1 Pro & \shortstack{35.0 [13.6, 56.2]\\35.0 [13.6, 56.2]} & \shortstack{30.0 [9.1, 52.6]\\30.0 [9.1, 52.6]} & \shortstack{30.0 [9.1, 52.6]\\30.0 [9.1, 52.6]} \\
Claude Opus 5.5 & \shortstack{30.0 [9.1, 53.3]\\25.0 [7.7, 44.4]} & \shortstack{25.0 [5.0, 47.4]\\20.0 [4.8, 37.5]} & \shortstack{20.0 [4.8, 37.5]\\15.0 [0.0, 33.3]} \\
GPT-6-astra & \shortstack{100.0 [100.0, 100.0]\\100.0 [100.0, 100.0]} & \shortstack{100.0 [100.0, 100.0]\\100.0 [100.0, 100.0]} & \shortstack{90.0 [73.7, 100.0]\\90.0 [73.7, 100.0]} \\
\bottomrule
\end{tabular}
\end{table}

\begin{table}[t]
\centering\small
\setlength{\tabcolsep}{5pt}
\caption{Gap width tolerance sensitivity on 20 tasks per model. Each cell gives A (top) and T (bottom), as percentages with 95\% whole-scene bootstrap intervals.}
\label{tab:tolerance_gap}
\begin{tabular}{@{}lccc@{}}
\toprule
Model & $1.00$ A / T & $0.75$ A / T & $0.50$ A / T \\
\midrule
GPT-5.6-sol & \shortstack{10.0 [0.0, 21.6]\\10.0 [0.0, 21.6]} & \shortstack{10.0 [0.0, 21.6]\\10.0 [0.0, 21.6]} & \shortstack{10.0 [0.0, 21.6]\\10.0 [0.0, 21.6]} \\
GPT-6.1-sol & \shortstack{60.0 [38.5, 76.9]\\60.0 [38.5, 76.9]} & \shortstack{50.0 [26.7, 72.7]\\50.0 [26.7, 72.7]} & \shortstack{45.0 [25.0, 71.4]\\45.0 [25.0, 71.4]} \\
DeepSeek-V4.1-Flash & \shortstack{5.0 [0.0, 20.0]\\5.0 [0.0, 20.0]} & \shortstack{5.0 [0.0, 20.0]\\5.0 [0.0, 20.0]} & \shortstack{5.0 [0.0, 20.0]\\5.0 [0.0, 20.0]} \\
Claude Sonnet 5.5 & \shortstack{10.0 [0.0, 29.4]\\10.0 [0.0, 29.4]} & \shortstack{10.0 [0.0, 29.4]\\10.0 [0.0, 29.4]} & \shortstack{10.0 [0.0, 29.4]\\10.0 [0.0, 29.4]} \\
Kimi K3.1 Pro & \shortstack{15.0 [0.0, 26.3]\\15.0 [0.0, 26.3]} & \shortstack{15.0 [0.0, 26.3]\\15.0 [0.0, 26.3]} & \shortstack{15.0 [0.0, 26.3]\\15.0 [0.0, 26.3]} \\
Claude Opus 5.5 & \shortstack{10.0 [0.0, 22.2]\\10.0 [0.0, 22.2]} & \shortstack{10.0 [0.0, 22.2]\\10.0 [0.0, 22.2]} & \shortstack{5.0 [0.0, 10.8]\\5.0 [0.0, 10.8]} \\
GPT-6-astra & \shortstack{75.0 [46.2, 92.6]\\75.0 [46.2, 92.6]} & \shortstack{70.0 [38.5, 89.5]\\70.0 [38.5, 89.5]} & \shortstack{60.0 [30.8, 77.8]\\60.0 [30.8, 77.8]} \\
\bottomrule
\end{tabular}
\end{table}

\begin{table}[t]
\centering\small
\setlength{\tabcolsep}{4pt}
\caption{All reversals of originally strict pairwise A/T ordering in the three numerical task types. Differences and paired whole-scene bootstrap intervals are percentage points. Intervals describe uncertainty in the new ordering; ties splitting are excluded.}
\label{tab:tolerance_rank_changes}
\begin{tabular}{@{}llccrr@{}}
\toprule
Task & Pair (first minus second) & Metric & Scale & Original gap & New gap [95\% CI] \\
\midrule
Object distance & Claude Sonnet 5.5 / Kimi K3.1 Pro & A & 0.50 & -15.0 & 10.0 [-7.5, 36.4] \\
Object distance & Claude Sonnet 5.5 / Claude Opus 5.5 & A & 0.50 & -30.0 & 5.0 [0.0, 13.3] \\
Object distance & GPT-6.1-sol / Claude Opus 5.5 & T & 0.75 & -5.0 & 15.0 [-9.5, 36.8] \\
Object distance & GPT-6.1-sol / Claude Opus 5.5 & T & 0.50 & -5.0 & 20.0 [-9.5, 50.0] \\
Room area & GPT-5.6-sol / Kimi K3.1 Pro & A & 0.50 & 20.0 & -5.0 [-33.3, 25.0] \\
Room area & DeepSeek-V4.1-Flash / Claude Sonnet 5.5 & A & 0.75 & -5.0 & 10.0 [-9.5, 29.4] \\
Room area & DeepSeek-V4.1-Flash / Claude Sonnet 5.5 & A & 0.50 & -5.0 & 15.0 [0.0, 31.6] \\
Room area & GPT-5.6-sol / Kimi K3.1 Pro & T & 0.50 & 10.0 & -15.0 [-42.1, 13.3] \\
\bottomrule
\end{tabular}
\end{table}

\subsection{Action Cost}
\label{sec:appendix_resources}

Failed episodes also consume actions. We therefore report mean actions over all episodes, mean actions among strict task successes, and total actions divided by strict task successes. The last measure includes failure costs. All 250 episodes per model have recorded high-level action counts. GPT-5.6-sol uses fewer actions on strict task successes than GPT-6.1-sol (40.96 versus 46.27), yet its total cost per success is higher (386.00 versus 108.60). Reliable completion can therefore matter more than shortening successful episodes.

\begin{table}[t]
\centering\small
\caption{Action cost across seven models. All 250 episodes per model have known action counts. Brackets show whole-scene bootstrap percentiles. The last column includes actions spent on failures.}
\label{tab:action_cost_intervals}
\begin{tabularx}{\textwidth}{Ylll}
\toprule
Model & Mean actions & \shortstack{Mean actions\\on T-passing episodes} & \shortstack{Total actions\\per T-passing episode} \\
\midrule
GPT-5.6-sol & 74.11 [59.29, 90.28] & 40.96 [27.03, 57.52] & 386.00 [261.78, 595.86] \\
GPT-6.1-sol & 50.82 [41.08, 61.85] & 46.27 [38.66, 54.43] & 108.60 [83.42, 142.45] \\
DeepSeek-V4.1-Flash & 66.27 [56.33, 76.99] & 35.54 [25.19, 47.55] & 637.23 [401.94, 1108.01] \\
Claude Sonnet 5.5 & 60.30 [47.43, 73.66] & 44.44 [35.99, 54.97] & 190.84 [130.96, 276.21] \\
Kimi K3.1 Pro & 79.10 [64.75, 94.11] & 56.85 [45.44, 67.97] & 278.54 [201.08, 392.86] \\
Claude Opus 5.5 & 53.18 [41.77, 65.82] & 43.62 [36.11, 52.92] & 132.96 [94.67, 184.30] \\
GPT-6-astra & 43.62 [34.96, 53.74] & 43.61 [35.88, 52.84] & 63.03 [48.84, 82.03] \\
\bottomrule
\end{tabularx}
\end{table}

\section{Visual and Behavioral Evidence Review}
\label{sec:appendix_visual}

\begin{figure}[t]
\centering
\includegraphics[width=\linewidth]{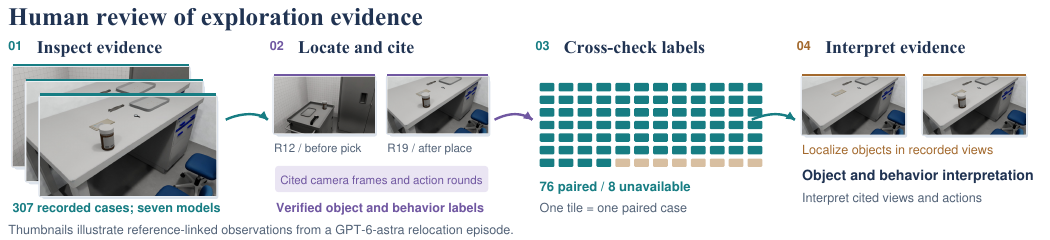}
\caption{Human review of exploration evidence across seven evaluated models. Reference images, exploration frames, and action feedback support manually checked object-localization and behavioral judgments. Tiles show paired-judgment availability for 84 recorded cases. Camera thumbnails illustrate the inputs using a GPT-6-astra relocation episode.}
\label{fig:visual_review_pipeline}
\end{figure}

\subsection{Human Review Protocol}
\label{sec:visual_review_protocol}
The 307-case review examines which objects can be localized and which behaviors occur in the recorded trajectories. This differs from the 200-episode audit of individual N/S conditions in Appendix~\ref{sec:ns_human_audit}. The 307 cases are sampled from the 1,750 VLM episodes, grouped by model, task specification, and strict task success. Each stratum contains up to two recorded cases, with weights reflecting the episodes it represents. All 307 cases have usable primary labels.

We examine original reference images, camera observations, action sequences, and feedback. Evidence packets contain up to six camera frames and feedback excerpts, covering the trajectory and its first and last object interactions. We verify cited object locations and behavior against these records.

\paragraph{Human Evaluation and Ethics.}
The human evaluations involve simulator-based task completion, numerical estimation from rendered views, and annotation of model trajectories. We do not seek formal ethics review because we consider these simulator-based evaluations low risk and collect no identifying personal information. All participants provide informed consent for participation and publication of aggregate results, and we report only these aggregate results.

\subsection{Human Audit of N and S}
\label{sec:ns_human_audit}
For direct comparison with automatic N/S scores, the separate 200-episode audit covers all seven models, 11 task types, and passing and failing N/S outcomes. It applies the task-specific rules in Appendix~\ref{sec:appendix_tasks}. With model identity, final answers, and automatic scores hidden, one annotator checks whether task-relevant objects appear in the delivered views and whether the recorded exploration and interactions meet each applicable N/S condition. Image, sensor, and event records support the checks, and each judgment cites its frame or event.

N and S are ratable in 196 and 194 episodes, respectively. Table~\ref{tab:ns_audit_counts} reports the raw comparison with automatic scores. Table~\ref{tab:ns_audit_rates} gives unweighted agreement, false-positive rates, and false-negative rates, with the single-annotator judgments as the reference. Agreement is 97.4\% for N and 93.3\% for S. This close agreement supports automatic checks of the task-required observations and interactions in N and the additional corroboration in S. Answer correctness is measured independently.

\begin{table}[t]
\centering\small
\setlength{\tabcolsep}{3pt}
\caption{Automatic checks against human judgments in 200 episodes. TP/TN agree; FP is an automatic pass with a human failure, and FN is the reverse.}
\label{tab:ns_audit_counts}
\begin{tabular}{@{}lrrrrrr@{}}
\toprule
Check & TP & FP & FN & TN & Ratable & Unknown \\
\midrule
N & 109 & 2 & 3 & 82 & 196 & 4 \\
S & 82 & 5 & 8 & 99 & 194 & 6 \\
\bottomrule
\end{tabular}
\end{table}

\begin{table}[t]
\centering\small
\setlength{\tabcolsep}{4pt}
\caption{Unweighted rule-audit rates: agreement $(\mathrm{TP}+\mathrm{TN})/n$, FP rate $\mathrm{FP}/(\mathrm{FP}+\mathrm{TN})$, and FN rate $\mathrm{FN}/(\mathrm{FN}+\mathrm{TP})$.}
\label{tab:ns_audit_rates}
\begin{tabular}{@{}lrrrr@{}}
\toprule
Check & Ratable & Agreement & FP rate & FN rate \\
\midrule
N & 196 & 97.4\% & 2.4\% & 2.7\% \\
S & 194 & 93.3\% & 4.8\% & 8.9\% \\
\bottomrule
\end{tabular}
\end{table}

\subsection{Object Localizability and Task-Required Evidence}
Four categories distinguish insufficient evidence, a referent that cannot be localized in the supplied views, ambiguous localization, and clear localization. A clear label requires a relevant referent, camera-frame ID, and normalized bounding box. It establishes that at least one relevant object or region can be localized in the recorded views.

\begin{table}[t]
\centering\small
\caption{Human-verified object localizability in recorded exploration views across seven models. Clear labels require a camera-frame bounding box for at least one relevant object or region. Weights reflect episode counts within model, task specification, and strict task success strata.}
\label{tab:visual_localizability}
\begin{tabularx}{\textwidth}{Yrrl}
\toprule
Model & Cases & Clear labels & Weighted clear (\%) [scene percentiles] \\
\midrule
GPT-5.6-sol & 43 & 32 & 61.80 [47.16, 77.75] \\
GPT-6.1-sol & 46 & 39 & 88.80 [78.89, 96.70] \\
DeepSeek-V4.1-Flash & 40 & 32 & 76.40 [58.64, 93.88] \\
Claude Sonnet 5.5 & 46 & 41 & 85.40 [70.92, 96.35] \\
Kimi K3.1 Pro & 46 & 36 & 70.80 [53.08, 88.92] \\
Claude Opus 5.5 & 45 & 40 & 89.80 [80.70, 97.67] \\
GPT-6-astra & 41 & 35 & 83.40 [69.72, 95.65] \\
\bottomrule
\end{tabularx}
\end{table}

Within the 307-case review sample, weighted clear localizability is 79.5\% overall and 71.2\% among its 35 cases with A=1 and N=0. These 35 cases are a subset of the 206 gaps categorized in Appendix~\ref{sec:appendix_gap_attribution}. A stricter check requires a localization for every supplied non-option reference. This holds in 148 of 281 eligible cases, with weighted prevalence 45.3\%. Reference roles follow their caption format, excluding option and encoded-choice labels. The manually confirmed localizations show that identifiable targets can coexist with unmet N conditions, such as observing every counted instance or covering the required rooms. S additionally checks task-specific corroboration, including geometric measurements.

\subsection{Consistency of Recorded Judgments}
We manually verify two sets of recorded localization judgments for 84 cases. Of these, 76 have usable pairs and eight have unavailable pairs. Category agreement is 85.5\%, with Cohen's $\kappa=0.595$. Assigning agreement or disagreement to all unavailable pairs gives a sensitivity range of 77.4--86.9\%.

Both recorded judgments label 57 cases clearly localizable. They identify a common frame--referent pair in 55, and corresponding boxes reach an intersection-over-union of at least 0.5 in 54. These manually checked comparisons describe consistency in category labels and object localization.

We also check the recorded image controls for exact image matching, blank-image handling, and simple object counting. Both sets of judgments meet the predefined control criteria. A camera-withholding control retains reference images and actions while removing exploration frames. Clear labels fall from 20 of 24 original judgments to zero without the camera images. This confirms that these localization judgments depend on the exploration images.

\subsection{Observable Behavior and Skill Constraints}
We manually check overlapping labels for difficult views, possible instance confusion, repeated search, explicit navigation or interaction failures, metric estimates without recorded measurement support, and incomplete task-required coverage. Every label requires a frame or action citation. An episode can receive several labels.

We identify explicit constraint reports in the same 307 reviewed cases using fixed text rules. Table~\ref{tab:feedback_constraints} reports unweighted episode and event counts, and weighted prevalence among T-failing and T-passing episodes. Positive reports can occur in episodes that later succeed.

\begin{table}[t]
\centering\small
\caption{Constraint reports in the 307 reviewed cases. Episode and event counts are unweighted; percentages are weighted prevalence conditional on T failure or success. Constraints can overlap and occur before successful recovery. Brackets show whole-scene bootstrap percentiles.}
\label{tab:feedback_constraints}
\begin{tabularx}{\textwidth}{Xrrll}
\toprule
Feedback constraint & Episodes & Events & T failures (\%) & T successes (\%) \\
\midrule
Obstacle-related drive feedback & 255 & 3608 & 82.48 [72.60, 91.82] & 80.37 [67.61, 90.85] \\
Ground skill applied to a raised point & 45 & 86 & 15.93 [9.44, 22.79] & 7.00 [3.14, 11.74] \\
Object skill applied to a ground point & 19 & 25 & 8.32 [3.53, 14.32] & 5.29 [1.26, 10.33] \\
Arm target on floor, wall, or ceiling & 14 & 25 & 9.55 [4.15, 15.06] & 0.49 [0.00, 1.21] \\
No reachable arm pose reported & 26 & 71 & 15.80 [8.52, 23.05] & 4.32 [1.53, 7.24] \\
Sensor feedback explicitly inconclusive & 36 & 74 & 17.96 [10.96, 25.01] & 10.10 [3.80, 17.37] \\
\bottomrule
\end{tabularx}
\end{table}

Obstacle reports are common in both T failures and successes, whereas arm targets on room surfaces and unreachable arm poses are more prevalent among failures. These patterns identify physically valid interaction targets and reachable approaches as useful checks alongside feedback-guided recovery.

\subsection{Statistical Analysis}
Localizability and behavioral estimates use the recorded stratum weights. Their ranges are the 2.5th and 97.5th percentiles of 10,000 whole-scene resamples. The N/S rule audit reports direct unweighted counts and rates.

\section{Evidence-Based Model Profiles}
\label{sec:appendix_bottlenecks}
Using the episode scores analyzed in Appendix~\ref{sec:posthoc_diagnostics}, we compare each model's spatial-answer errors after observation checks pass and its relocation failure stages. The human review in Appendix~\ref{sec:appendix_visual} helps interpret the objects and behavior behind these patterns.

\subsection{Conditional Errors and Relocation Stages}
We examine spatial-answer errors after N passes, with additional analyses after S passes, and locate failures along the pick-and-place sequence. Recorded scores, interaction events, camera views, and additional archived observations connect these measurements to each episode.

For G1--G4, $P(A=0\mid N=1)$ measures incorrect outcomes among episodes passing task-required evidence checks (Table~\ref{tab:bottleneck_outputs}). For G2 and G3, conditioning on both N and S examines errors after stronger corroboration also passes (Table~\ref{tab:bottleneck_stronger}). Each cell reports errors over episodes passing the stated N/S checks. G4 combines relative direction and path integration.

Using the milestones in Appendix~\ref{sec:appendix_relocation_milestones}, we assign episodes with incorrect relocation outcomes (A=0) to three stages: before correct pick, after correct pick but before correct place, and after correct place with the final goal unmet. The stages partition A failures within the same 210 relocation episodes. Uncorrected wrong-object pickup is an additional label that can overlap these stages.

\begin{table}[t]
\centering
\footnotesize
\setlength{\tabcolsep}{4pt}
\caption{Incorrect outcomes among N-passing episodes. Cells give errors/episodes in grounding (G1), aggregation (G2), geometry (G3), and reference-frame integration (G4).}
\label{tab:bottleneck_outputs}
\begin{tabularx}{\columnwidth}{Ycccc}
\toprule
Model & G1 & G2 & G3 & G4 \\
\midrule
GPT-5.6-sol & 11/33 & 14/19 & 18/31 & 3/9 \\
GPT-6.1-sol & 11/37 & 15/28 & 14/68 & 9/30 \\
DeepSeek-V4.1-Flash & 17/27 & 6/12 & 23/30 & 8/10 \\
Claude Sonnet 5.5 & 5/36 & 5/15 & 33/50 & 8/20 \\
Kimi K3.1 Pro & 9/37 & 20/28 & 30/50 & 6/17 \\
Claude Opus 5.5 & 5/37 & 9/24 & 35/62 & 7/24 \\
GPT-6-astra & 5/39 & 14/34 & 6/74 & 6/48 \\
\bottomrule
\end{tabularx}
\end{table}

\begin{table}[t]
\centering
\footnotesize
\setlength{\tabcolsep}{3pt}
\caption{Relocation failure stages (30 tasks/model). The three stage columns partition A failures. Pre-place follows correct pickup; post-place follows ordered correct placement. Wrong counts uncorrected wrong-object pickups and can overlap these stages.}
\label{tab:bottleneck_execution}
\begin{tabularx}{\columnwidth}{Yrrrrr}
\toprule
Model & Failed & Pre-pick & Pre-place & Post-place & Wrong \\
\midrule
GPT-5.6-sol & 28 & 21 & 6 & 1 & 20 \\
GPT-6.1-sol & 27 & 15 & 9 & 3 & 15 \\
\shortstack[l]{DeepSeek-V4.1-\\Flash} & 29 & 25 & 4 & 0 & 16 \\
Claude Sonnet 5.5 & 21 & 12 & 5 & 4 & 10 \\
Kimi K3.1 Pro & 26 & 16 & 8 & 2 & 15 \\
Claude Opus 5.5 & 20 & 13 & 5 & 2 & 12 \\
GPT-6-astra & 20 & 14 & 1 & 5 & 14 \\
\bottomrule
\end{tabularx}
\end{table}

\begin{table}[t]
\centering
\footnotesize
\caption{Incorrect counting (G2) and geometry (G3) answers among episodes passing both N and S. Cells give errors over episodes passing N and S.}
\label{tab:bottleneck_stronger}
\begin{tabular}{lcc}
\toprule
Model & G2, N and S & G3, N and S \\
\midrule
GPT-5.6-sol & 9/14 & 15/26 \\
GPT-6.1-sol & 14/22 & 12/63 \\
DeepSeek-V4.1-Flash & 3/6 & 17/21 \\
Claude Sonnet 5.5 & 5/10 & 26/40 \\
Kimi K3.1 Pro & 16/23 & 24/43 \\
Claude Opus 5.5 & 8/18 & 27/54 \\
GPT-6-astra & 12/27 & 4/68 \\
\bottomrule
\end{tabular}
\end{table}

\subsection{Model Profiles}
The profiles highlight each model's main spatial-answer and relocation difficulties. Conditional fractions refer to N-passing episodes unless both N and S are stated. Relocation failure counts refer to A=0; T additionally requires the task's observation and interaction checks.

\paragraph{GPT-5.6-sol.}
Among N-passing episodes, GPT-5.6-sol answers 14/19 aggregation questions and 18/31 geometric questions incorrectly. Of its 28 relocation failures, 21 occur before correct pickup, and 20 include an uncorrected wrong-object pick. Reliable counting across views and verification of the picked object's identity are therefore priorities.

\paragraph{GPT-6.1-sol.}
GPT-6.1-sol has 15/28 aggregation errors and 14/68 geometric errors after N passes. Its relocation failures split 15/9/3 across the three stages. The nine failures between correct pickup and placement identify reliable transfer as a priority alongside cross-view counting.

\paragraph{DeepSeek-V4.1-Flash.}
DeepSeek-V4.1-Flash has 17/27 grounding errors, 23/30 geometric errors, and 8/10 reference-frame errors after N passes. Of 29 relocation failures, 25 occur before correct pickup. Only five of the 30 tasks reach correct pickup. Reference-frame judgment and intended-target pickup are clear improvement targets.

\paragraph{Claude Sonnet 5.5.}
Claude Sonnet 5.5 has 33/50 geometric errors after N passes, compared with 5/36 grounding errors. After both N and S pass, geometric errors remain at 26/40, including 9/9 gap-width questions. Relocation failures split 12/5/4 across the stages. Gap-width reasoning and target verification before pickup are therefore priorities.

\paragraph{Kimi K3.1 Pro.}
Kimi K3.1 Pro has 20/28 aggregation errors and 30/50 geometric errors after N passes. The corresponding N-and-S fractions are 16/23 and 24/43. Relocation failures split 16/8/2, with 15 uncorrected wrong-object pickups. Cross-view counting, geometric precision, and verification of the selected object are therefore priorities.

\paragraph{Claude Opus 5.5.}
Claude Opus 5.5 has 35/62 geometric errors after N passes, compared with 5/37 grounding errors. Geometric errors remain at 27/54 after both checks pass, including 11/13 gap-width questions. Relocation failures split 13/5/2, with 12 uncorrected wrong-object pickups. Geometric precision and correct target pickup are therefore priorities.

\paragraph{GPT-6-astra.}
GPT-6-astra has 14/34 aggregation errors and 6/74 geometric errors after N passes. After N and S pass, these fractions are 12/27 and 4/68. Fourteen of its 20 relocation failures occur before correct pickup, all with uncorrected wrong-object picks. Fifteen episodes reach correct placement, but five still end without the accepted goal. With stronger geometry, cross-view counting, target verification, and final-state checks after placement become prominent improvement targets.

\subsection{Alignment with Related Error Analyses}
EmbodiedBench~\cite{yang2025embodiedbench} analyzes perception, reasoning, and planning errors along the agent pipeline. We similarly locate failures in recorded interaction stages. MMSI-Bench~\cite{yang2025mmsi} separates grounding, overlap matching and scene reconstruction, situation-transformation reasoning, and spatial logic. Our conditional error rates connect spatial answers to the observations verified during an episode. OpenEQA~\cite{majumdar2024openeqa} distinguishes episodic memory from active information gathering. Our history diagnostics examine how acquired views contribute to later answers.

\subsection{Reproducibility}
Episode records link task identities, scores, and interaction stages. Appendix~\ref{sec:appendix_terminal_history} compares retrieval from each model's history with target-view supplementation from other trajectories.

\section{GT Retrieval and View Supplementation}

\label{sec:appendix_terminal_history}

We compare recovering task-relevant observations from each model's own exploration trajectory with supplying target observations beyond that trajectory. Both settings preserve the task, public references, accumulated exploration context, and answer instruction, modifying only the observations supplied before the final answer. Baseline outcomes are reused from each model's original episode. Table~\ref{tab:terminal_history_shared} reports per-model results and their pooled totals.

\begin{table}[t]
\centering
\small
\setlength{\tabcolsep}{6pt}
\caption{Correct answers/total (accuracy) on 139 questions per model. Both interventions retain each model's own exploration context. View supplementation adds target observations beyond that trajectory. Original outcomes are reused.}
\label{tab:terminal_history_shared}
\begin{tabular}{lccc}
\toprule
Model & Original episodes & GT retrieval & \shortstack{GT view\\supplementation} \\
\midrule
GPT-5.6-sol & 38/139 (27.3\%) & 45/139 (32.4\%) & 58/139 (41.7\%) \\
GPT-6.1-sol & 81/139 (58.3\%) & 85/139 (61.2\%) & 89/139 (64.0\%) \\
DeepSeek-V4.1-Flash & 22/139 (15.8\%) & 34/139 (24.5\%) & 46/139 (33.1\%) \\
Claude Sonnet 5.5 & 59/139 (42.4\%) & 64/139 (46.0\%) & 74/139 (53.2\%) \\
Kimi K3.1 Pro & 62/139 (44.6\%) & 72/139 (51.8\%) & 101/139 (72.7\%) \\
Claude Opus 5.5 & 74/139 (53.2\%) & 76/139 (54.7\%) & 91/139 (65.5\%) \\
GPT-6-astra & 104/139 (74.8\%) & 105/139 (75.5\%) & 109/139 (78.4\%) \\
\midrule
All seven models & 440/973 (45.2\%) & 481/973 (49.4\%) & 568/973 (58.4\%) \\
\bottomrule
\end{tabular}
\end{table}

\paragraph{Common questions and model-specific histories.}

Eligible question-answering tasks require active exploration and have recoverable final-answer inputs, recorded task-target views, and generated GT trajectory observations covering the task targets. The common set contains 139 tasks in 26 scenes: 32 in G1, 34 in G2, 52 in G3, and 21 relative-direction tasks in G4. Each model retains its own original exploration history in both settings. The diagnostic changes only the final judgment call and measures answer accuracy.

\paragraph{Archived retrieval.}

We construct GT retrieval inputs by retrieving all frames containing task-relevant objects from each model's archived trajectory before its original final answer. These frames are added to the retrieved observations in the final-answer context. The rest of the original context is retained, so this setting recovers evidence the model already encounters during exploration. The retrieved image count follows the available relevant frames in that trajectory.

\paragraph{GT view supplementation.}

We construct GT view supplementation inputs from target views recorded in other trajectories of the same task. We add these views to each model's own final-answer context, including observations of targets it has not seen. GT retrieval recovers views already present in that model's history, while supplementation can fill gaps in its target observations.

GT retrieval gives 481/973 correct answers (49.4\%), compared with 440/973 (45.2\%) in original episodes. GT view supplementation gives 568/973 (58.4\%), an 8.9-percentage-point improvement over GT retrieval. All seven models improve in both comparisons. Retrieval gains range from 0.7 to 8.6 percentage points across models, and supplementation adds 2.9--20.9 points. These gains show that additional task-relevant observations can improve final spatial judgments. Supplementation supplies views beyond the original trajectory; view selection and image counts follow the construction described above.

\section{GT Trajectory and Spatial Sensors}

\label{sec:appendix_oracle_evidence}

\subsection{GT Trajectory Inputs}

GT trajectory judgment replaces exploration context with generated GT trajectory observations, retaining the task question and public reference images. Models answer without further exploration. The set contains the same 139 active-exploration tasks used for the historical-image comparisons.

\begin{table}[t]
\centering
\small
\setlength{\tabcolsep}{6pt}
\caption{GT trajectory judgments on 139 questions per model. Both conditions use identical RGB views.}
\label{tab:oracle_accuracy}
\begin{tabular}{lcc}
\toprule
Model & GT trajectory RGB & GT trajectory + sensors \\
\midrule
GPT-5.6-sol & 79/139 (56.8\%) & 112/139 (80.6\%) \\
GPT-6.1-sol & 97/139 (69.8\%) & 128/139 (92.1\%) \\
DeepSeek-V4.1-Flash & 59/139 (42.4\%) & 70/139 (50.4\%) \\
Claude Sonnet 5.5 & 86/139 (61.9\%) & 96/139 (69.1\%) \\
Kimi K3.1 Pro & 102/139 (73.4\%) & 113/139 (81.3\%) \\
Claude Opus 5.5 & 98/139 (70.5\%) & 119/139 (85.6\%) \\
GPT-6-astra & 110/139 (79.1\%) & 117/139 (84.2\%) \\
\midrule
All seven models & 631/973 (64.9\%) & 755/973 (77.6\%) \\
\bottomrule
\end{tabular}
\end{table}

\begin{table}[t]
\centering
\small
\setlength{\tabcolsep}{6pt}
\caption{Shared diagnostic coverage across seven models. G1 includes object-room questions, container contents, and room comparison. G2 covers object and room counting, G3 geometric measurements, and G4 relative direction. Every condition uses this same question set.}
\label{tab:sensor_packet}
\begin{tabular}{lrr}
\toprule
Task group & Questions & \shortstack{Evaluations\\per condition} \\
\midrule
G1 & 32 & 224 \\
G2 & 34 & 238 \\
G3 & 52 & 364 \\
G4 & 21 & 147 \\
Total & 139 & 973 \\
\bottomrule
\end{tabular}
\end{table}

The 36 counting and room-comparison questions use 16 distinct scene RGB observations. These cover all task targets and include at least two background views. Remaining views are chosen by deterministic scene sampling without a fixed quota per target. The 17 container-contents questions use 40 slots, applying the same view-selection rule to all public options. The other 86 questions use 12 slots.

We construct GT trajectory observations from scene geometry and task-target annotations, then provide annotation-free RGB views. We check that the observations cover the task targets. Frame IDs do not indicate target identity or selection roles, and the views are presented in a shuffled order. The model is told that views can repeat objects and must be matched to unique instances. Model-visible materials contain the task, public references, and scene observations. Target masks, instance counts, and correctness labels remain in private selection and scoring records.

\subsection{Spatial Sensor Inputs}

\label{sec:appendix_sensor_packet}

The sensor condition keeps the generated GT trajectory RGB observations and their order identical. It adds aligned radial-depth samples, camera positions, forward directions, and camera calibration. Radial depth measures distance from the camera in meters. The camera uses a Z-up convention with equal horizontal and vertical focal lengths. These readings supply metric scale and coordinates across views.

Sensor inputs use depth-grid images except for the 17 container-contents questions answered by the three GPT models, which receive the same depth samples as text matrices. Thus, 922 model--question evaluations use image grids and 51 use text matrices.

\subsection{Answers and Comparisons}

Each model answers once per question and input condition. The first returned model response is retained. The content rule accepts a standalone answer command or one unique command on the final nonempty line after prose. Multiple commands, nonstring arguments, and trailing text are rejected.

Every condition contains 973 model--question evaluations: seven models on the same 139 questions. Correct-answer rates use all evaluations as the denominator. Unusable answers count as incorrect. Original outcomes are reused from each model's recorded exploration episode. GT trajectory inputs replace that exploration context, while the sensor contrast adds depth, pose, and calibration jointly to identical RGB observations. Table~\ref{tab:oracle_accuracy} reports per-model results and their pooled totals.

GT trajectory RGB gives 631/973 correct answers (64.9\%), 6.5 percentage points above GT view supplementation. This comparison jointly replaces the target views and the exploration context. Adding spatial sensors to identical RGB inputs gives 755/973 (77.6\%), a further gain of 12.7 percentage points. The sensor comparison supports providing physical scale and common coordinates together with target views.

Together, these results show that target observations beyond the original history and joint geometric feedback improve final spatial judgments. GT trajectory RGB still leaves 342/973 incorrect or unusable answers, compared with 218/973 after adding sensors. These remaining errors suggest pairing richer inputs with more reliable cross-view instance matching and task-specific use of geometric measurements. The diagnostics evaluate one final answer on active-exploration tasks; the full benchmark separately checks whether complete episodes satisfy T.

\section{Evaluation Configuration}
\label{sec:appendix_configuration}
Appendix~\ref{sec:appendix_tasks} describes the observation and action flow within an episode. Here we collect the complete inference, input, and skill configuration for all seven models in the main 250-task evaluation. All evaluations use a temperature of 0 and top-p of 1, with reasoning effort set to the highest level supported by each model. Table~\ref{tab:shared_configuration} summarizes the shared input and interaction limits. Task-specific reference images and initial states follow the benchmark task files.

\begin{table}[t]
\centering\small
\setlength{\tabcolsep}{4pt}
\caption{Shared main-evaluation observation and interaction limits. Reference images and task-defined budgets are resolved separately for each episode.}
\label{tab:shared_configuration}
\begin{tabularx}{\textwidth}{@{}L{0.23\textwidth}Y@{}}
\toprule
Parameter & Main-evaluation configuration shared by all seven models \\
\midrule
Camera images & RGB JPEG images, 768 pixels wide and 432 pixels high (quality 90). \\
Images per decision & At most 50 images, including recent views, reference images, retrieved images, intermediate views, and the current path visualization. \\
Visual history & Most recent 20 visual rounds; textual dialogue is retained. Selected intermediate views join this history. \\
Intermediate views & At most eight from the preceding action, sampled at 1\,m of translation or 30° of rotation. Shown at the first decision in the round, with optional retention. \\
Reference images & Task-specific isolated object or surface references. Category exemplars support counting; encoded choices identify answer options. Reference files follow each recorded model--task configuration. \\
RGB, depth, and pose & RGB observations, action and gripper feedback, and a cumulative odometry path visualization. Depth is available to the evaluator; the main agent receives the visual and proprioceptive feedback described here. \\
Retrieval & At most two retrieval passes and 12 archived images in total per decision, with their original text. Empty or invalid requests consume one retrieval opportunity. \\
Action budget & 25 times the task's physical-room count (25--325 actions), charged per high-level action decision. Up to eight additional rounds accommodate invalid outputs. \\
\bottomrule
\end{tabularx}
\end{table}

\paragraph{Embodiment and skill execution.}
The evaluated humanoid uses fixed-pose body geometry approximately 0.53\,m long, 0.36\,m wide, and 1.32\,m high. Navigation planning uses a 0.20\,m clearance radius. Execution checks oriented body geometry with a 0.01\,m margin and stops at collisions. Manipulation checks horizontal reach to the selected surface or nearest object boundary, using a nominal 1.5\,m limit and a 0.05\,m gate tolerance. Image-point commands refer to the latest delivered camera image and are mapped to the selected object or 3D surface. The platform's high-level skills validate these geometry and access conditions before applying the requested grasp, release, or articulation transition. Outcome verification separately checks the designated object's identity and destination, while S adds post-placement support verification.

\paragraph{Output validation and transport retries.}
The parser checks the action command and arguments before execution. Invalid outputs receive parsing feedback and produce no environment action. Eight cumulative unparsable outputs end an episode. The transport client allows at most ten attempts per call, with a 120-second request timeout and bounded backoff for transient failures. When a provider explicitly rejects an optional request parameter, the client removes that parameter before retrying. A nonempty response returns to the parser and is not regenerated; unavailable or unusable final answers count as incorrect.

\paragraph{Single-answer input diagnostics.}
The four GT settings retain the question and public references and follow the input protocols in the preceding diagnostic sections. They obtain one final response per model and question. Sensors augment identical GT trajectory RGB views with radial depth, camera pose, and calibration. Original episode outcomes are reused as the baseline.

\ifdefined\roamutftokens\else

\ifdefined\pdfgentounicode
  \input glyphtounicode.tex
  \pdfgentounicode=1
\fi
\makeatletter
\newcommand{\roamutftokens}[1]{%
  \begingroup
  \let\do\@makeother\dospecials
  \catcode32=10
  \endlinechar=-1
  \expandafter\scantokens\expandafter{\detokenize{#1}}%
  \endgroup}
\makeatother
\fi
\subsection{Example System Prompt}

\label{sec:recorded_prompt}

The following example shows the system-prompt format for a surface-contents task. Task instructions, reference captions, and action budgets vary by task. The final paragraphs specify retrieval, intermediate-frame retention, and cumulative odometry. The accompanying code provides the prompt templates and per-round observation messages.

\begingroup\small\ttfamily\raggedright

\setlength{\parindent}{0pt}

\roamutftokens{You are the decision-making brain of a household robot in a simulated house.}\par\smallskip

\roamutftokens{Each round you receive three things: a photograph from the robot's own camera, one
line of text saying whether your hand is full, and a report of what the robot's
own body measured while carrying out your last action. You issue exactly one
action per round. There is no description of this house, no list of what is in it,
and nothing anywhere that names anything -- everything you come to know about the
place you will have worked out from pictures.}\par\smallskip

\roamutftokens{- It is 768x432 pixels with a wide lens, taken from the robot's eye height. Small
  or distant things are only a few pixels wide and are easy to mistake for
  something else, so treat a distant identification as a guess until you have
  walked closer.
- Where a thing sits in the frame is not a hint about your action, it **is** your
  action: every skill takes a pixel coordinate read off this image. Look at the
  picture before every single move.
- You are shown the 20 most recent photographs, not just the newest one. Use
  them together with the body report: the report says what the robot *did*, and
  the difference between consecutive frames says what that achieved. Where the
  report stops short -- and it says so when it does -- the pictures are all you
  have.
- `holding:` says `(something)` or `(nothing)`, and that is the only fact about
  the world's state you are ever given in words. It will never say what you are
  carrying: a gripper measures force and finger position, not names.}\par\smallskip

\roamutftokens{You are given no text description of this house. Not the rooms it has, not the
furniture in it, not the name of a single object, and no list that fills in as
you explore. The one line of text you get each round is `holding:`, and it says
only whether your hand is full or empty -- never what is in it.}\par\smallskip

\roamutftokens{Your senses are the photograph and what the robot can feel in its own body. Your
only memory is this conversation: every THOUGHT line you write is kept and shown
back to you on every later round, while the older photographs are dropped. So the
notes you take are the only durable record you will have -- when you see that the
kitchen is through the left-hand doorway, write that down, because the picture
proving it will be gone. If you pick something up and it matters later what it
was, write that down too: the `holding:` line will not remind you.}\par\smallskip

\roamutftokens{Your task:
Go to the nightstand. Which of the following objects are resting on it? Select every option that applies. Answer only with the option IDs.
Options:
- OBJ_01: coffee table book
- OBJ_02: hardcover novel
- OBJ_03: drinking glass
- OBJ_04: dish drying rack
- OBJ_05: stovetop kettle
- OBJ_06: bedside table lamp
- OBJ_07: knife block
- OBJ_08: scented candle}\par\smallskip

\roamutftokens{9 images are attached above, showing what the task is about:
  Image 1: the nightstand this task refers to -- this exact object
  Image 2: Option OBJ_01 — coffee table book; category reference image
  Image 3: Option OBJ_02 — hardcover novel; category reference image
  Image 4: Option OBJ_03 — drinking glass; category reference image
  Image 5: Option OBJ_04 — dish drying rack; category reference image
  Image 6: Option OBJ_05 — stovetop kettle; category reference image
  Image 7: Option OBJ_06 — bedside table lamp; category reference image
  Image 8: Option OBJ_07 — knife block; category reference image
  Image 9: Option OBJ_08 — scented candle; category reference image
Each is rendered on a plain white background -- the object has been cut out of its surroundings, so the white says nothing about where it is. These are reference pictures, not what you see now.}\par\smallskip

\roamutftokens{Skills available to you this run (and only these):
  go_to_object_at_pixel(u, v)
  go_to_ground_at_pixel(u, v)
  move(dist_m)
  strafe(dist_m)
  turn(deg)
  look(deg)
  pick(u, v)
  place(u, v)
  open(u, v)
  close(u, v)
  answer("...")   -- give your final answer to a question, in English
  done()          -- declare the task finished}\par\smallskip

\roamutftokens{Rules that matter:
- You refer to things by **pointing at them in the photograph**, and there is no
  other way to refer to anything. Every skill below takes `u, v`: `u` is the
  column counting from the **left** edge, `v` is the row counting from the **top**
  edge, both in pixels, so `u` is 0-767 and `v` is 0-431 on this 768x432
  frame. So (384, 216) is the centre of the view, (384, 401) is the
  floor just in front of your feet, and (384, 30) is up near the ceiling.
  `open(384, 216)` opens whatever is at that point.
- Aim for the **middle** of the thing you mean, not its edge. An edge pixel
  usually belongs to whatever is behind it, and you will act on that instead.
- Objects and ground are **separate skills**, and choosing the wrong one fails
  rather than guessing: `go_to_object_at_pixel(u, v)` walks up to a *thing* and
  stops within reach of it, while `go_to_ground_at_pixel(u, v)` walks onto a patch
  of *floor* and stands there. The robot's depth camera can work out whether the
  point you gave lies on the floor plane or stands above it, so when you pick the
  wrong skill it says which -- and then the thing to change is the skill, not the
  pixel.
- Nothing names anything for you. What you are pointing at is entirely your own
  reading of the image, with no second opinion to check it against, so a
  coordinate that looks like a mug handle may be a spoon or a shadow. When it
  matters, walk closer and look again before you act.
- To put something somewhere: pick(u, v) on the thing, then walk to the
  destination, then place(u, v) pointing at the destination -- `place` points at
  where it goes, the held object is implied. The `holding:` line tells you whether
  your hand is full but never what is in it, and a held object is also carried in
  front of the camera, in the lower middle of the frame.
- `go_to_object_at_pixel(u, v)` only reaches what is on screen at this
  instant -- that is what pointing means. Something you saw two rooms ago is
  memory, not a target: there is no pixel you can give for it. If what you
  want is not in the frame, turn, look or walk until it is; that search is
  your job, not the skill's.
- `go_to_ground_at_pixel(u, v)` is how you cross a room. Point it at floor
  as far away as you can still see clearly -- through a doorway, at the far
  wall's base -- and you will walk there. It needs no object in view, so it
  is the right tool for getting into a room you cannot see into yet.
- For measured motion, positive move is forward, positive strafe is right,
  positive turn is left and positive look is up. Negative values reverse
  each direction.
- You have a budget of 100 actions. Rooms are not all visible from where you
  stand; turning and walking to look around is expected and costs budget.
- **After each action you are told what the robot's own body measured, and
  nothing else.** Each line opens with the instrument it came from, and these are
  the ones you will see: `Odometry:` -- the wheels counting how far the base
  rolled and how far it turned; `Head encoder:` -- whether the camera's pitch
  joint moved or hit the end of its travel; `Gripper:` -- whether anything is
  between the fingers; `Wrist force sensor:` -- whether the thing the hand pushed
  on actually moved under it; `Depth camera:` and `Local map:` -- what the range
  image and the floor plan made of the point you pointed at; `Arm:` -- whether a
  reachable arm pose exists from where the base is standing. The label tells you
  what kind of claim you are reading.
- Those lines are measurements, not judgements, and they are complete. "The
  fingers closed and came back empty" is not a refusal with the reason left out
  -- it is everything the robot felt. Nothing will ever tell you "that is not
  something you can pick up", or "you cannot reach that from here", or what any
  object is called, because no sensor on the robot measures those things.
- Placement feedback ends with "read the new photograph", and means it literally.
  When you put something down, the hand knows it let go and nothing on the robot
  knows where the object came to rest. You have to see this for yourself.
- An action that achieved nothing leaves the view **exactly** as it was, so two
  identical photographs in a row mean your last action did nothing at all -- and
  the thing to change is the action, not the number of times you try it.
- You are still told when your own message could not be used: a reply this harness
  could not parse, a call that is not on the skill list, a coordinate off the edge
  of the frame. Those are about what you wrote rather than about the house, and
  they arrive before the robot moves at all.}\par\smallskip

\roamutftokens{
Unified memory and action decision. Recent observation images remain visible; older rounds have an indexed summary, chosen action and original execution receipt. All previous original observation/reply text is archived and retrievable. Summaries are uncertain notes, not ground truth. Choose EITHER retrieve information OR one normal action in each reply. After retrieval you see its results and make the SAME decision again. Retrieval does not spend physical action steps; at most two retrieval requests are allowed per action round, including empty or invalid requests. No more flag or separate selector is needed. If the information is sufficient, select the action directly. Avoid repeatedly retrieving the same landmarks without a concrete unresolved question.
For retrieval reply:
THOUGHT: <brief information gap>
RETRIEVE: {"retrieve": [image_ids], "retrieve_text": [integer_rounds]}
For an action reply:
MEMORY: <at most 100 words of observed facts, distinctive objects/rooms, orientation, counting/de-duplication evidence and unresolved uncertainty>
THOUGHT: <brief reason>
ACTION: <one call from the skills above>
Never output RETRIEVE and ACTION in the same reply. The current action has NOT executed; only summarize its outcome after the next execution receipt. All historical images have historical coordinates. Use ONLY the CURRENT FINAL observation for action pixels.
C intermediate memory: the first decision previews ALL candidate intermediate frames from the previous action. On that first reply optionally add KEEP: [candidate_frame_ids] before other headers, whether retrieving or acting. Default KEEP is []. Keep only key views; unselected candidates are discarded from later requests and cannot be retrieved. Kept views enter recent history and the archive alongside the final observation. Keep selection is final for this round. Image IDs use R#.F and R#.I#; text IDs use integer round numbers.
CONTENTS CHECK: inspect the entire named surface or container from enough views to cover it. Include only objects directly on or in that target, keep distinct instances, and exclude nearby objects on another surface. Before answer(), check this evidence once; retrieve a relevant old view only when a specific part of the target remains unresolved.}\par\smallskip

\roamutftokens{Self-trajectory sensor: every decision includes one cumulative top-down odometry plot through the current pose. It contains only the robot's own x/z/yaw history, metric grid and step order; it contains no floor plan, room boundary, room label, object location or task truth. Use it to reason about where you have moved and turned. The CURRENT FINAL camera image remains the last image and is the only source for action pixel coordinates.}\par\smallskip

\endgroup

\endgroup

\end{document}